\documentclass[10pt,twocolumn,letterpaper]{article}

\usepackage{wacv}
\usepackage{times}
\usepackage{epsfig}
\usepackage{graphicx}
\usepackage{amsmath}
\usepackage{amssymb}
\usepackage{booktabs}
\usepackage{caption}
\usepackage{array}
\usepackage{tabularx}
\usepackage[dvipsnames]{xcolor}
\usepackage[skip=0cm,list=true]{subcaption}
\usepackage{url}
\usepackage{multicol}
\usepackage{makecell}
\usepackage{comment}
\usepackage{tikz}
\usepackage{tabularx}
\usepackage[accsupp]{axessibility}

\usepackage{cite}

\usepackage{microtype}
\usepackage{colortbl}

\usepackage{soul}
\usepackage{todonotes}

{\vspace{-\topsep}\begin{itemize}\itemsep1pt \parskip0pt \parsep0pt}
{\end{itemize}\vspace{-\topsep}}

\definecolor{gold(bg)}{rgb}{0.87, 0.79, 0.32}
\definecolor{gold(metallic)}{rgb}{0.83, 0.69, 0.22}
\definecolor{gold(web)(golden)}{rgb}{1.0, 0.84, 0.0}

\definecolor{palesilver}{rgb}{0.79, 0.75, 0.73}
\definecolor{silver}{rgb}{0.75, 0.75, 0.75}
\definecolor{lightslategray}{rgb}{0.47, 0.53, 0.6}

\definecolor{bronze(bg)}{rgb}{0.9, 0.6, 0.3}
\definecolor{bronze}{rgb}{0.8, 0.5, 0.2}

\definecolor{goldL}{HTML}{FBF2D2}
\definecolor{silverL}{HTML}{DDDDDD}
\definecolor{bronzeL}{HTML}{EED2B8}

\definecolor{goldD}{HTML}{D9AE13}
\definecolor{silverD}{HTML}{909090}
\definecolor{bronzeD}{HTML}{9A5F26}

\newcommand*\circledd[3]{\tikz[baseline=(char.base)]{
            \node[shape=circle,fill=#1,draw=#2,inner sep=1pt] (char) {\small{#3}};}}

\newcommand{\mfirst}[1]{%
    {#1\raisebox{0.8pt}{\circledd{goldL}{goldD}{1}}}%
}
\newcommand{\msecond}[1]{%
    {#1\raisebox{0.8pt}{\circledd{silverL}{silverD}{2}}}%
}
\newcommand{\mthird}[1]{%
    {#1\raisebox{0.8pt}{\circledd{bronzeL}{bronzeD}{3}}}%
}

\definecolor{mygray}{gray}{0.8}
\newcommand{\oldfirst}[1]{%
    {#1\raisebox{0.8pt}{\circledd{mygray}{mygray}{1}}}%
}
\newcommand{\oldsecond}[1]{%
    {#1\raisebox{0.8pt}{\circledd{mygray}{mygray}{2}}}%
}
\newcommand{\oldthird}[1]{%
    {#1\raisebox{0.8pt}{\circledd{mygray}{mygray}{3}}}%
}

\newcommand{\medal}[3]{\tikz[baseline=(char.base)]{\node[rounded corners=2pt,fill=#1,draw=#2,inner sep=1.5pt] (char) {#3};}}

\newcommand{\bm}[2]{
    \ifcase#1\or
      {\medal{goldL}{goldD}{\textbf{#2}}}
    \or 
      {\medal{silverL}{silverD}{#2}}
    \or 
      {\medal{bronzeL}{bronzeD}{#2}}
    \else 
      #2
    \fi\ignorespaces
}

\newcolumntype{L}[1]{>{\raggedright\let\newline\\\arraybackslash\hspace{0pt}}b{#1}}
\newcolumntype{C}[1]{>{\centering\let\newline\\\arraybackslash\hspace{0pt}}b{#1}}

\newcommand{\rankn}[1]{({\small\##1})}

\def\wacvPaperID{****} 

\wacvfinalcopy 

\ifwacvfinal
\usepackage[breaklinks=true,bookmarks=false,hidelinks=True]{hyperref}
\else
\usepackage[pagebackref=true,breaklinks=true,colorlinks,bookmarks=false]{hyperref}
\fi

\begin{document}

\title{3\textsuperscript{rd} Workshop on Maritime Computer Vision (MaCVi) 2025: Challenge Results}

\author{Benjamin Kiefer$^{0,1}$, Lojze Žust$^2$, Jon Muhovič$^2$, Matej Kristan$^2$, Janez Perš$^2$, 
Matija Teršek$^3$,\\ Uma Mudenagudi$^6$, Chaitra Desai$^6$, Arnold Wiliem$^{4,5}$, Marten Kreis$^1$, Nikhil Akalwadi$^6$, \\ Yitong Quan$^1$,
Zhiqiang Zhong$^7$, Zhe Zhang$^7$, Sujie Liu$^7$, Xuran Chen$^7$, Yang Yang$^7$, \\ Matej Fabijanić$^8$, Fausto Ferreira$^8$, Seongju Lee$^9$, Junseok Lee$^9$, Kyoobin Lee$^9$,\\ Shanliang Yao$^{17}$, Runwei Guan$^{11}$, Xiaoyu Huang$^{10}$, Yi Ni$^{12}$, Himanshu Kumar$^{13}$, Yuan Feng$^{14}$, \\Yi-Ching Cheng$^{15}$, Tzu-Yu Lin$^{15}$, Chia-Ming Lee$^{15}$, Chih-Chung Hsu$^{15}$, Jannik Sheikh$^{16}$, \\Andreas Michel$^{16}$, Wolfgang Gross$^{16}$, Martin Weinmann$^{21}$, Josip Šarić$^{2}$, Yipeng Lin$^{7}$, Xiang Yang$^{7}$, \\Nan Jiang$^{18}$, Yutang Lu$^{19}$, Fei Feng$^{19}$, Ali Awad$^{20}$, Evan Lucas$^{20}$, Ashraf Saleem$^{20}$,\\ Ching-Heng Cheng$^{15}$, Yu-Fan Lin$^{15}$, Tzu-Yu Lin$^{15}$, Chih-Chung Hsu$^{15}$
{\tt\small  }
\and
$^0$LOOKOUT, 
$^1$University of Tuebingen, 
$^2$University of Ljubljana, 
$^3$Luxonis,
$^{4}$Shield AI,\\
$^{5}$Queensland University of Technology,
$^{6}$Center of Excellence in Visual Intelligence, \\KLE Technological University,
$^{7}$Nanjing University of Science and Technology,\\
$^{8}$University of Zagreb Faculty of Electrical Engineering and
Computing,
$^{9}$Gwangju Institute\\ of Science and Technology (GIST),
$^{10}$University of Liverpool, 
$^{11}$Hong Kong \\University of Science and Technology (Guangzhou),
$^{12}$Xi’an Jiaotong-Liverpool University,\\
$^{13}$Independent Researcher,
$^{14}$Dalian Maritime University, School of Marine Engineering,\\
$^{15}$National Cheng Kung University,
$^{16}$Fraunhofer IOSB,
$^{17}$Yancheng Institute of Technology,\\
$^{18}$Nanjing University,
$^{19}$Beijing University of Posts and Telecommunications,\\
$^{20}$Michigan Technological University
$^{21}$Karlsruhe Institute of Technology
}
\maketitle
\thispagestyle{empty}

\begin{abstract}
The 3$^{\text{rd}}$ Workshop on Maritime Computer Vision (MaCVi) 2025 
addresses maritime computer vision for Unmanned Surface Vehicles (USV) and underwater.

This report offers a comprehensive overview of the findings from the challenges. We provide both statistical and qualitative analyses, evaluating trends from over 700 submissions. All datasets, evaluation code, and the leaderboard are available to the public at \url{https://macvi.org/workshop/macvi25}.
\end{abstract}

\section{Introduction}

\begin{figure}[t]
\centering
\begin{subfigure}{.2\textwidth}
   \includegraphics[width=\textwidth,trim={0 0 30px 0},clip]{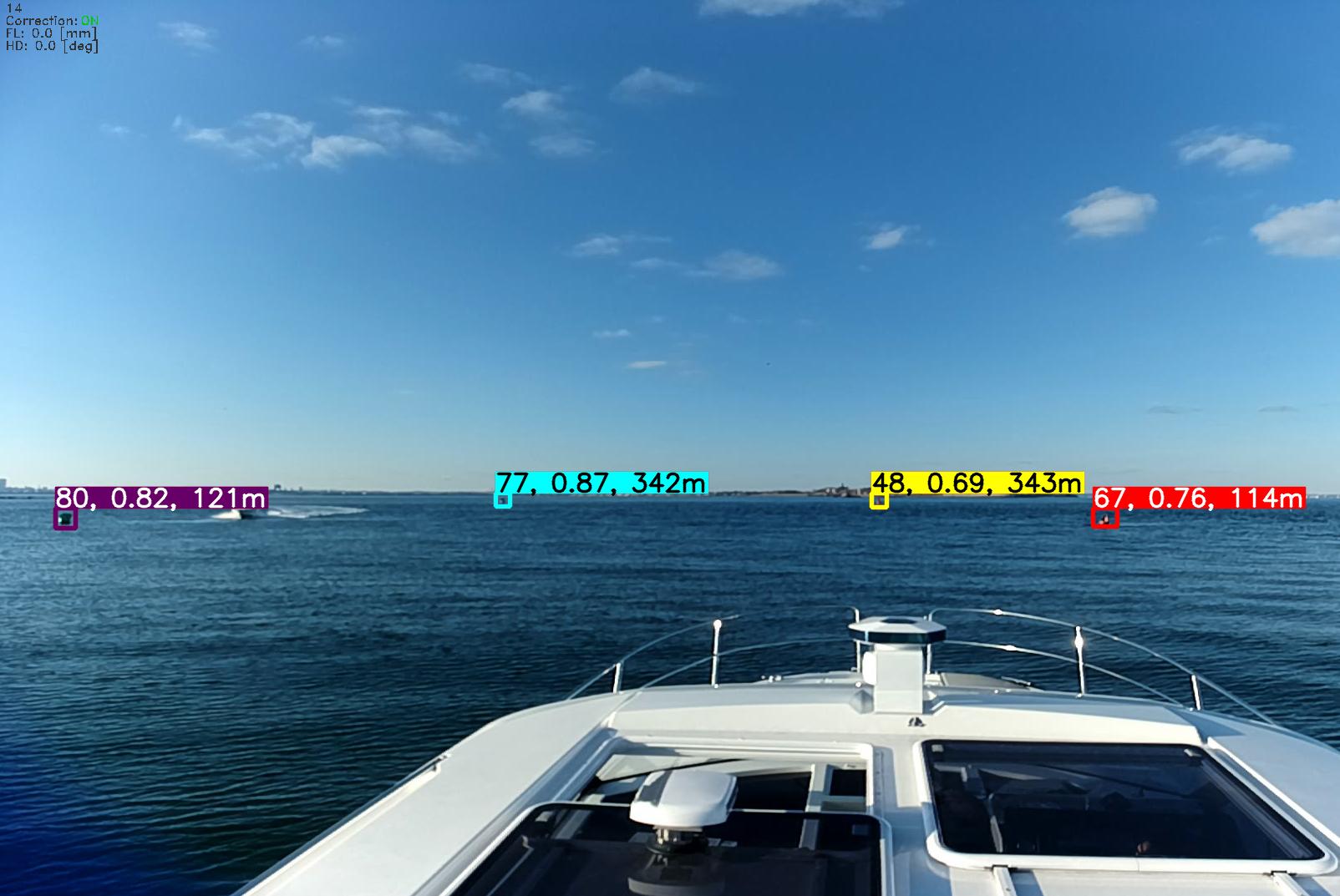}
   \caption{Distance Estimation}
   \label{fig:dist_estimation}
\end{subfigure}
\begin{subfigure}{.2\textwidth}
   \includegraphics[width=\textwidth]{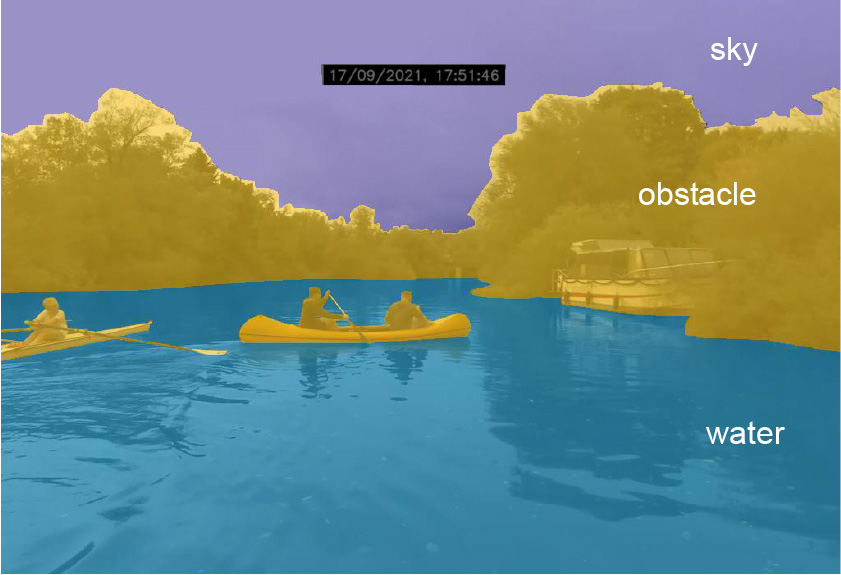}
   \caption{Obstacle Segmentation}
   \label{fig:obstacle_segmentation}
\end{subfigure}
\begin{subfigure}{.2\textwidth}
   \includegraphics[width=\textwidth]{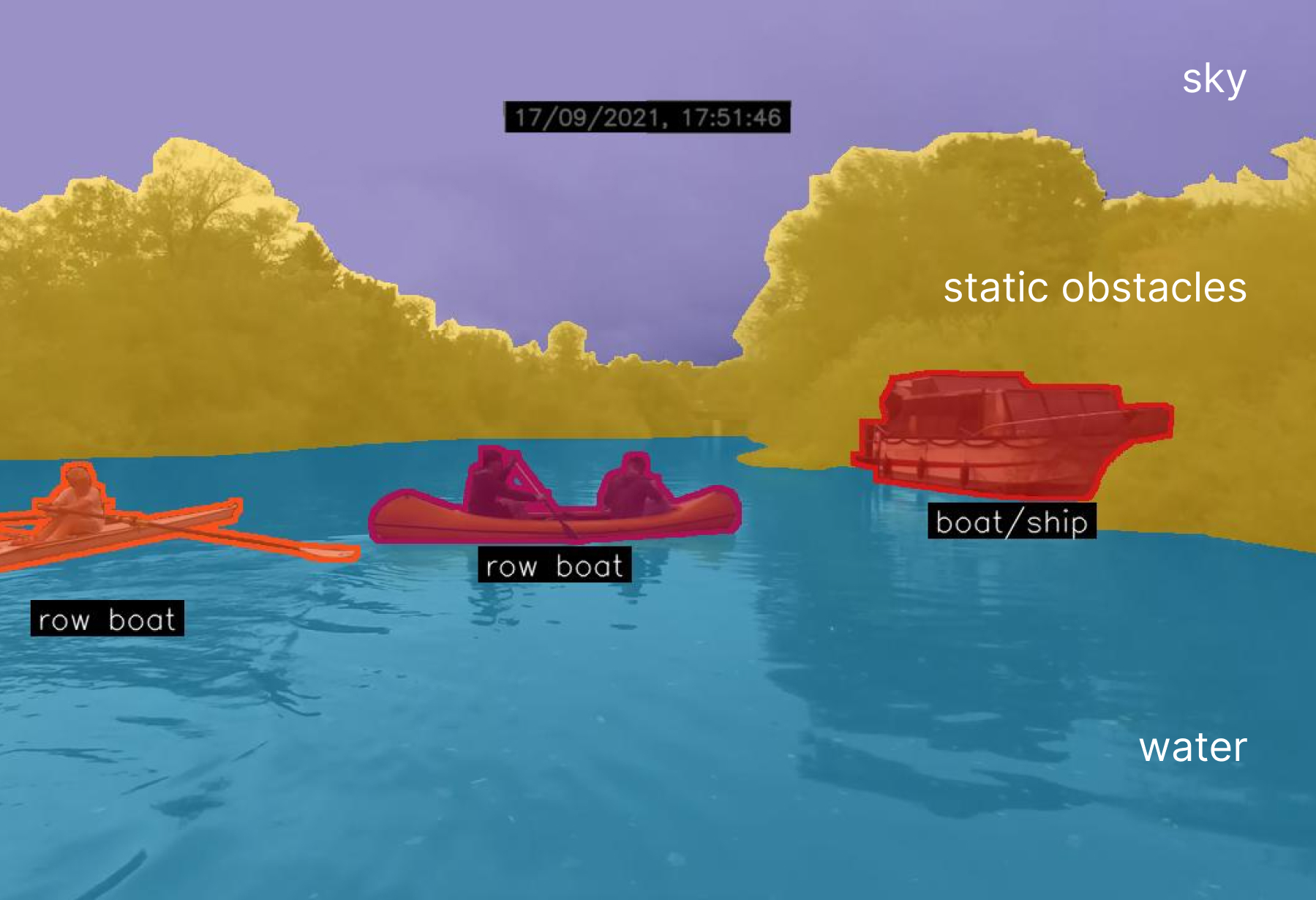}
   \caption{Panoptic Segmentation}
   \label{fig:panoptic_segmentation}
\end{subfigure}
\begin{subfigure}{.2\textwidth}
   \includegraphics[width=\textwidth,trim={0 0 211px 0},clip]{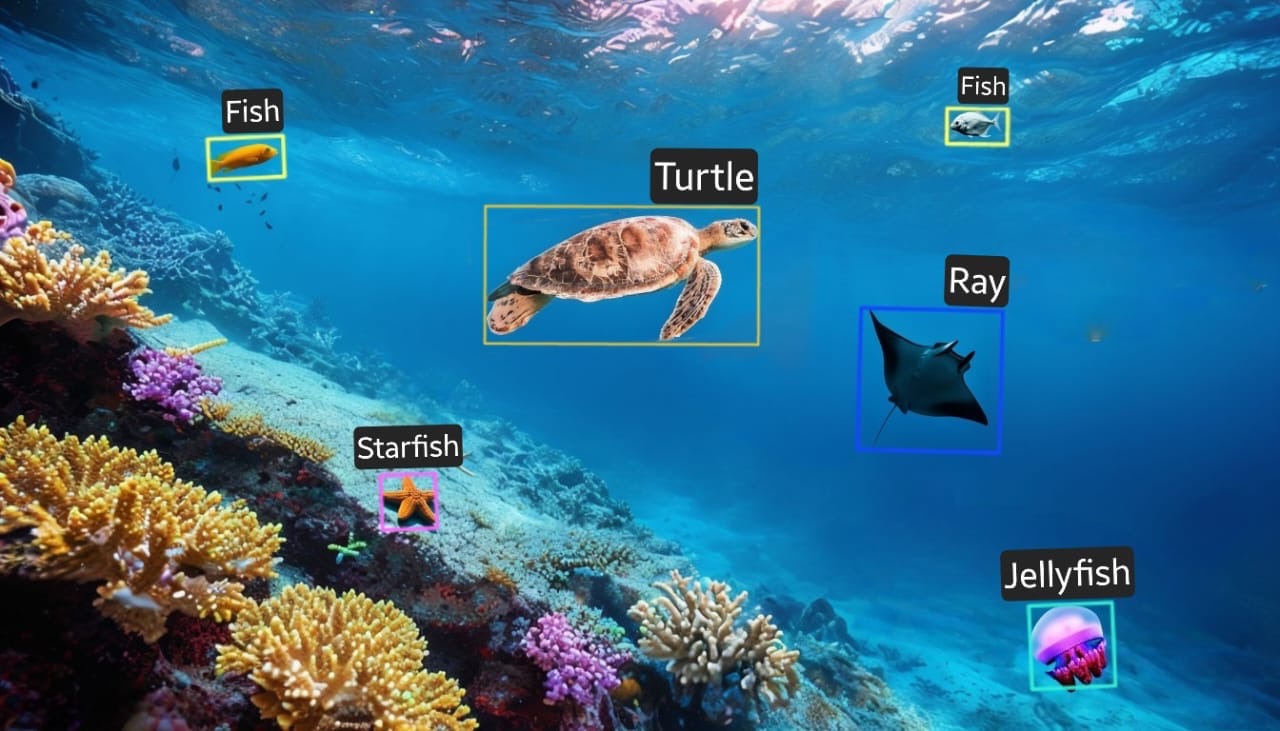}
   \caption{Image Restoration}
   \label{fig:marine_restoration}
\end{subfigure}
\caption{Overview of MaCVi2025 challenges. Challenges include USV-based (a) Distance Estimation, (b) Regular and Embedded Obstacle Segmentation, (c) Panoptic Segmentation, and UUV-based (d) Image Restoration. }
\label{fig:challenges_overview}
\end{figure}

Maritime environments, with their unique challenges such as dynamic lighting, reflections, and cluttered scenes, demand specialized computer vision techniques \cite{varga2022seadronessee,prasad2019object,kanjir2018vessel,gallego2019detection,kiefer2021leveraging}. Autonomous systems like Unmanned Surface Vehicles (USVs) and Underwater Vehicles (UUVs) rely heavily on robust vision algorithms to navigate, detect, and interpret complex surroundings \cite{Zust2022Learning,Bovcon2021,kiefer2025approximatesupervisedobjectdistance}. Addressing these challenges requires not only cutting-edge algorithms but also standardized benchmarks and a collaborative research ecosystem \cite{KristanPAMI2016,bovcon2019mastr,rsuigm,bloisi2014background}.

The 3\textsuperscript{rd} Workshop on Maritime Computer Vision (MaCVi 2025) builds on the momentum of the previous two iterations \cite{Kiefer_2023_WACV,Kiefer_2024_WACV} to foster innovation in maritime computer vision. This year, the workshop presented a suite of challenges, including distance estimation, (embedded) semantic \& panoptic segmentation, and image restoration. These tasks reflect advancements in dataset availability and evaluation protocols while emphasizing real-world deployment, including embedded hardware.

This report highlights the methodologies, results, and insights from the MaCVi 2025 challenges. The rest of the paper is structured as follows: Section~\ref{sec:protocol} details the general challenge protocols while Sections~\ref{sec:usvchallenges} and \ref{sec:uir-challenge} discuss the outcomes of individual tracks. All datasets, evaluation tools, and leaderboards are available at \url{https://macvi.org/workshop/macvi25}.

\section{Challenge Participation Protocol}
\label{sec:protocol}

As for the first two challenge iterations, all challenges followed a superset of rules and protocols outlined in this section while challenge specific rules are described in the respective subsections.

\subsection{Submission Process}
Participants could get the challenge rules and resources on the workshop webpages. There, we provided links to evaluation code, datasets and starter kits.

Then, participants submitted their models and predictions through the official evaluation server. General submission requirements included:
\begin{itemize}
    \item Predictions formatted according to standardized task-specific formats.
    \item Compliance with model submission guidelines, including export to ONNX format where applicable.
    \item A limit of 1-3 submission per day per challenge track.
\end{itemize}

\subsection{Evaluation Server and Leaderboard}
An online evaluation server provided automated scoring for all submissions. Test set evaluations were conducted on the server, with results displayed on public leaderboards. The leaderboards were frozen (results of each method visible only to its authors) a week prior to the results announcement to encourage competition and submissions without relying on the results from other teams.

\subsection{Timeline}
The timeline for all MaCVi 2025 challenges was:
\begin{itemize}
    \item \textbf{Submissions open:} 13th Sep 2024
    \item \textbf{Leaderboards frozen:} 12th Dec 2024
    \item \textbf{Submissions closed:} 19th Dec 2024
\end{itemize}

\subsection{Recognition and Rewards}
After additional manual review of the submissions, the top team(s) in each challenge were determined by the challenge-specific metrics and whether the performance was above the provided baselines. They then were invited to submit a short technical report detailing their methods. These reports will are included in the appendix below. Winners were also acknowledged during the workshop and featured on the official website. Additionally, there were prizes for some of the competitions as described on the workshop website.

\section{USV-based Perception Challenges}
\label{sec:usvchallenges}

\subsection{Approximate Supervised Distance Estimation}
\label{sec:usvdistance}
Distance estimation using monocular vision is a complex problem that has gained considerable attention in recent years due to its key role in fields such as autonomous driving and robotics\cite{masoumian2022mde}. This approach is equally crucial for maritime applications, where it can support essential tasks such as navigation and path planning for unmanned surface vehicles (USVs). State-of-the-art monocular depth estimation frameworks, such as MiDAS\cite{ranftl2020midas} and SPI-Depth\cite{lavreniuk2024spidepth}, show significant limitations for this type of application, particularly in terms of real-time performance and accuracy for smaller, distant objects. In addition, many of these approaches generate depth maps with values normalized between 0 and 1, rather than providing the metric distances required for practical use.\\

In the context of this challenge, participants are required to develop novel methods for estimating distances to maritime navigational aids while simultaneously detecting them in images. To facilitate this, we provide a dataset comprising approximately 3,000 labeled training samples, including single-class normalized bounding-box coordinates and the metric distance between the camera and the object. The ground truth distance is calculated using the Haversine formula, which computes the distance between the GNSS coordinates of the ship, where the camera is mounted, and the corresponding navigational buoys as mapped on a nautical chart. Additionally, participants are provided with a modified version of the YOLOv7 object detector\cite{wang2022yolov7}, similar to an approach developed by Vajgl \textit{et al.}\cite{vajgl_dist-yolo_2022}, where the predicted tensor for each anchor consisting of [$c_x$, $c_y$, $w$, $h$, $p_{obj}$, $p_{class\_1}$, ..., $p_{class\_n}$] is extended by a distance estimate.\\

\newcommand{\coloredsquareblue}{\tikz\shade[left color=black!80!blue, right color=blue] (0,0) rectangle (0.2,0.2);}
\newcommand{\coloredsquarered}{\tikz\shade[left color=black!80!red, right color=red] (0,0) rectangle (0.2,0.2);}
\begin{figure}[t]
\centering  
   \includegraphics[width=\linewidth]{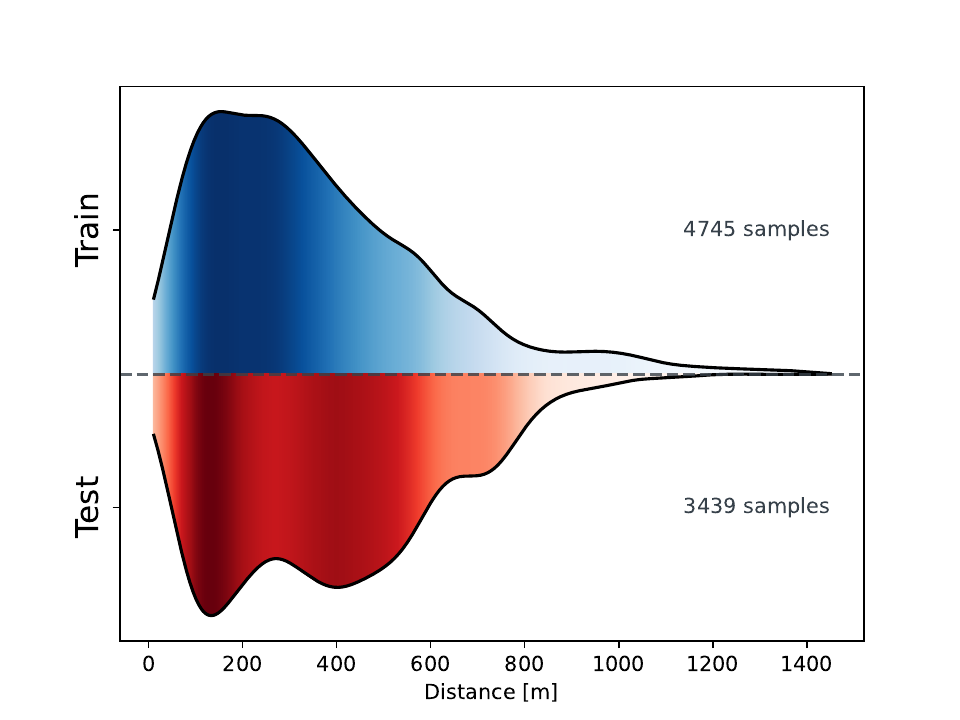}
\caption{Distribution of samples in the training (\protect\coloredsquareblue) and test (\protect\coloredsquarered) splits with respect to the ground truth distance}
\label{fig:dist_distribution}
\end{figure}

\begin{figure}[t]
\centering  
   \includegraphics[width=0.9\linewidth]{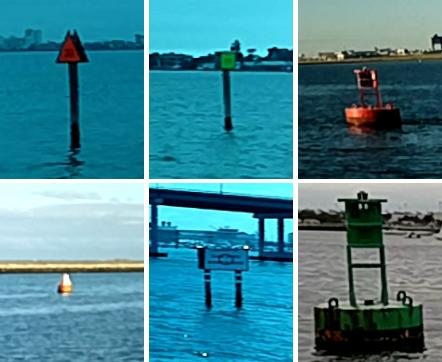}
\caption{Common buoy types in the supervised distance estimation dataset}
\label{fig:dist_buoytypes}
\end{figure}

The complete dataset includes 2,940 annotated images in the training split and 2,268 in the test split. A detailed distribution of individual objects in both splits, categorized by ground truth distance, is depicted in Figure \ref{fig:dist_distribution}. We chose to include challenging scenarios where buoys are distant and barely visible, aiming to assess the limits of the object detection models. Maritime navigational aids exhibit significant variation in size and appearance. Figure \ref{fig:dist_buoytypes} highlights the most common buoy types, predominantly comprising red and green channel markers.

\subsubsection{Evaluation Protocol}
To evaluate the methods submitted by participants, we utilize metrics that assess both object detection performance and distance estimation accuracy. For object detection performance, we employ well-established metrics, specifically $AP_{50}$ and $AP_{50:95}$. The $AP_{50}$ metric calculates the average precision across varying confidence thresholds with an \textit{IoU} threshold of $50\%$. In contrast, $AP_{50:95}$ averages the precision over a range of \textit{IoU} thresholds from $50\%$ to $95\%$ in increments of $5\%$.\\
To evaluate the deviation of distance estimates from ground truth values, we define the \textit{absolute error} ($\epsilon_{abs}$) and \textit{relative error} ($\epsilon_{rel}$) as follows:
\begin{align*}
    &\epsilon_{abs} = \frac{1}{n} \sum_{i=1}^n \left| \hat{d}_i - d_i \right|,\\
    &\epsilon_{rel} = \frac{1}{n} \sum_{i=1}^n \frac{c_i}{\sum_{j=1}^n c_j} \left| \frac{\hat{d}_i - d_i}{d_i} \right|,
\end{align*}
where $n$ is the total number of predictions, $\hat{d}_i$ denotes the predicted distance, $d_i$ represents the ground truth distance and $c_i$ is the predicted confidence. It is important to note that each distance prediction is weighted by the corresponding normalized confidence for $\epsilon_{rel}$. Furthermore, we include the relative distance error to penalize deviations between the prediction and ground truth more heavily for objects that are in closer proximity.\\
Finally, the combined quality measure $Q$ is defined as
\begin{align*}
    Q = AP_{50:95} \cdot (1 - \min(\epsilon_{rel}, 1)),
\end{align*}
incorporating both object detection and distance estimation performance.\\
\\
The submitted models must comply with the following guidelines for performance considerations.
\begin{itemize}
    \item Model size should not exceed 50 million parameters
    \item Models have to be exported to \textit{ONNX} format 
    \item The onnx export should not include \textit{NMS}
    \item Images of testset are rescaled to \textit{1024x1024}
\end{itemize}

\subsubsection{Submissions, Analysis and Trends}

\begin{table*}[ht!]
\caption{Comparison of the performance of the submitted methods and the organizer-provided baseline model (highlighted in gray) on the test split. Performance metrics include object detection measures: $AP_{50}$, $AP_{50-95}$, and distance estimation metrics: relative error ($\epsilon_{rel}$) and absolute error ($\epsilon_{abs}$). The overall quality score (\textbf{Q}) evaluates both object detection and distance estimation capabilities.}
\label{tab:usvdist-results}
\centering
\begin{tabular}{cllccccc}
\toprule
 Place &         Method &           Institution &          \bf{Q} $\downarrow$ &   $AP_{50}$ &         $AP_{50-95}$ &   $\epsilon_{rel}$&   $\epsilon_{abs}$\\
\midrule
\mfirst{} &     Data Enhance &      NJUST-KMG &            \bm1{0.2719} &          \bm1{0.742} &        \bm1{0.320} &    \bm2{0.149} &        \bm2{54.450} \\
\msecond{} &    Yolov7 Depth Widened & UNIZG FER &        \bm2{0.2554} &          \bm2{0.729} &        \bm2{0.299} &    \bm1{0.145} &        \bm1{51.170} \\
\midrule
-   & \color{gray} Baseline & \color{gray} MaCVi &          0.2260 &                0.676 &             0.274 &          0.175 &                59.608 \\
\bottomrule
\end{tabular}
\end{table*}

\begin{figure}[t]
\centering  
\includegraphics[width=\linewidth]{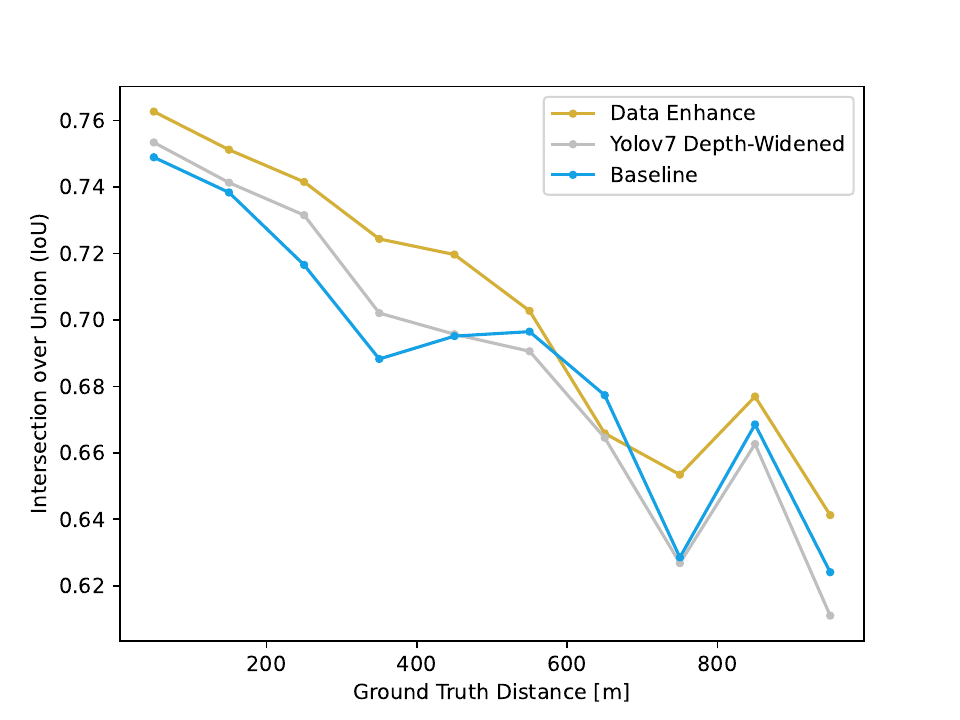}
\caption{Mean IoU over the distance to the objects with a step size of 100 meters for the top 2 submissions compared with the baseline model}
\label{fig:usvdist_iou_over_dist}
\end{figure}

\begin{figure*}[ht!]
\centering
\begin{subfigure}{.45\textwidth}
    \centering
    \includegraphics[width=\textwidth]{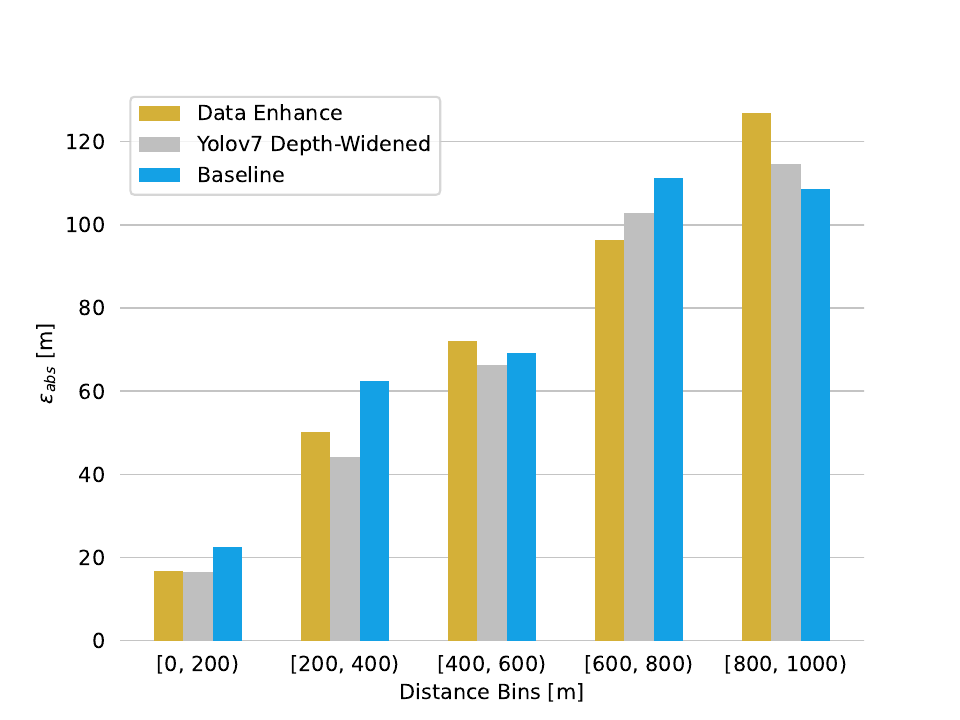}  
\end{subfigure}%
\begin{subfigure}{.45\textwidth}
    \centering
    \includegraphics[width=\textwidth]{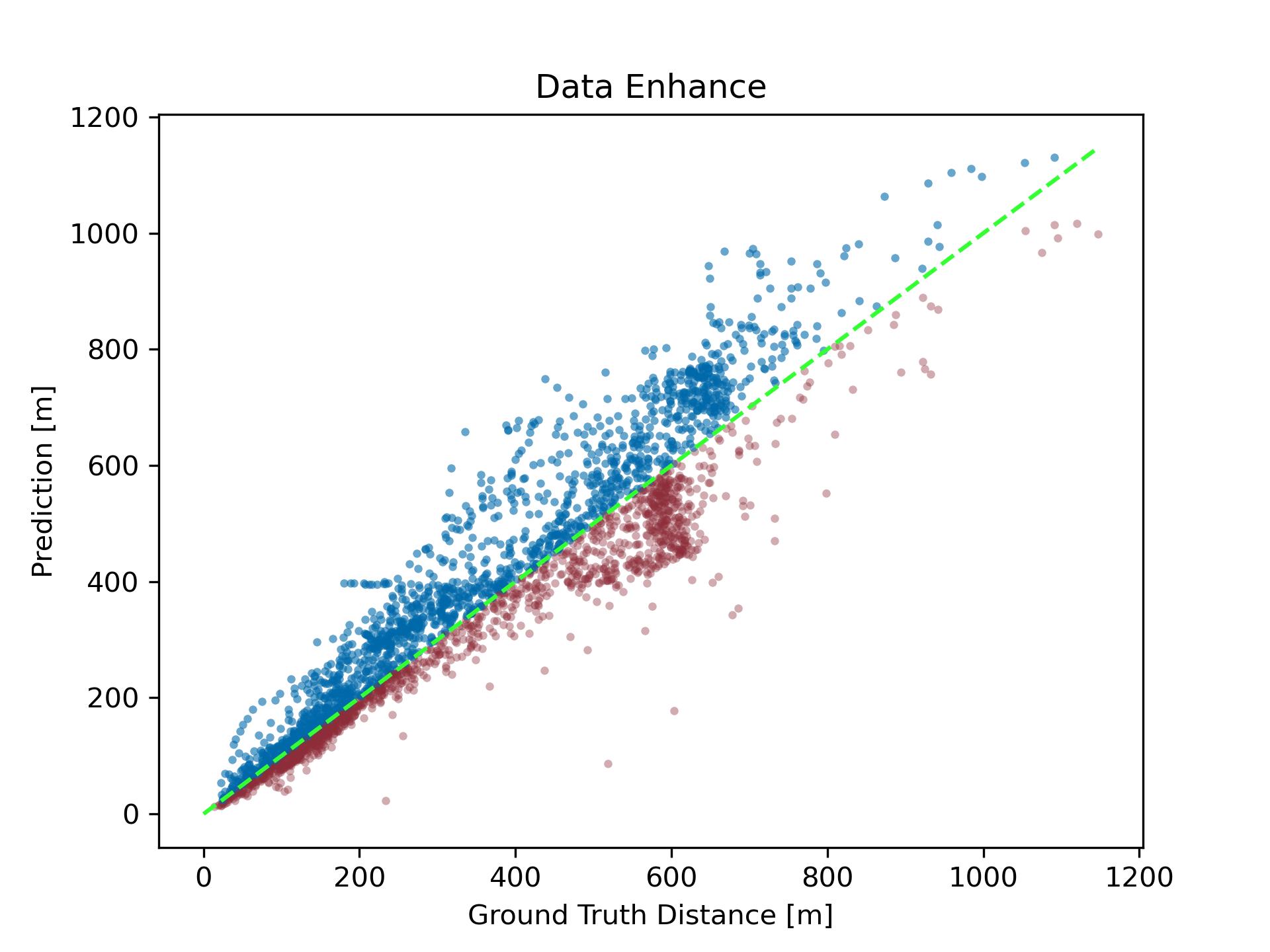}  
\end{subfigure}

\begin{subfigure}{.45\textwidth}
    \centering
    \includegraphics[width=\textwidth]{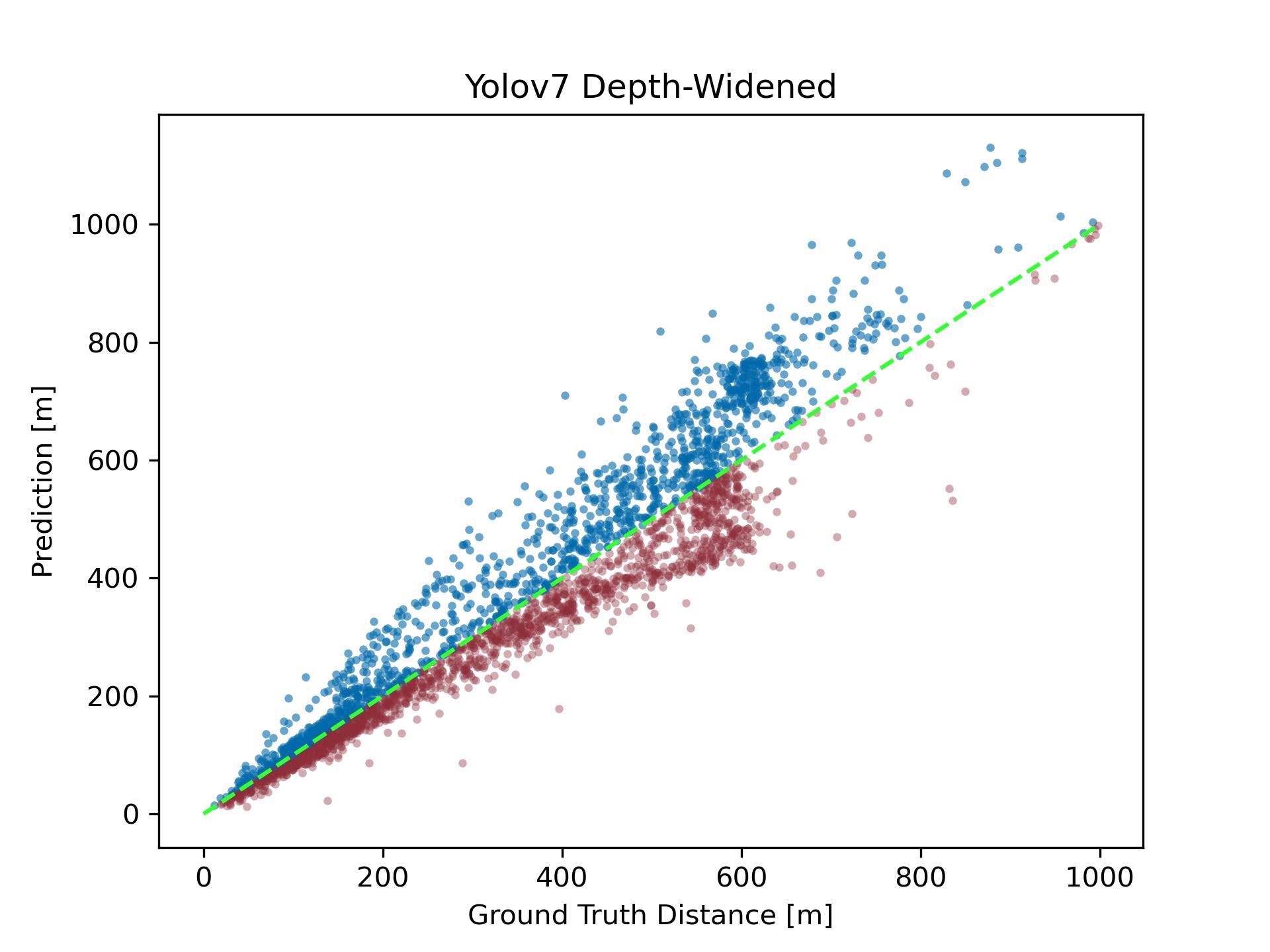}  
\end{subfigure}%
\begin{subfigure}{.45\textwidth}
    \centering
    \includegraphics[width=\textwidth]{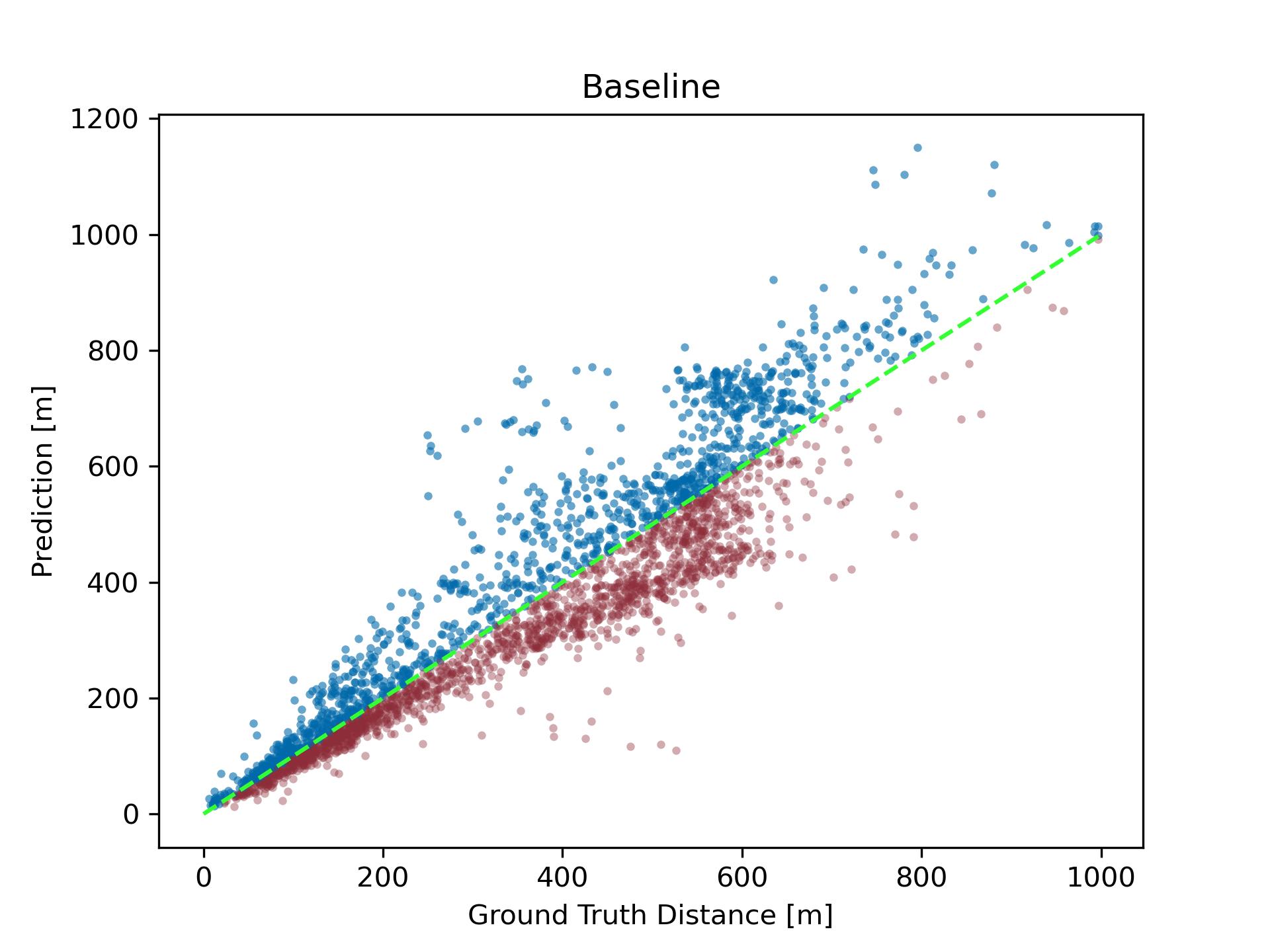}  
\end{subfigure}
\caption{Distance Estimation Results comparing the top two submissions, \textit{Data Enhance} and \textit{Yolov7 Depth Widened}, against the provided baseline model with the absolute error metric ($\epsilon_{abs}$).}
\label{fig:usvdist-disterror}
\end{figure*}

60 Submission from 6 different teams, including one baseline model from the MaCVi2025 committee \cite{kiefer2025approximatesupervisedobjectdistance}, were evaluated. Table \ref{tab:usvdist-results} displays the results of the challenge. Both top submissions outperform the baseline model in object detection performance and distance estimation metrics. However, the winning model is significantly better with regards to the object detection (AP) metrics, while the second-place submission excels in minimizing the distance estimation errors. In the following section the top two submissions are compared to the provided baseline model. Figure \ref{fig:usvdist-disterror} illustrates the distance estimation capabilities of the submitted models. The best-performing model for this task is \textit{Yolov7 Depth-Widened}, however both approaches are capable of significantly reducing the absolute error compared to the baseline, particularly for objects in closer proximity. This is mainly achieved by limiting the number of large outlier estimates and can likely be attributed to the design of the challenge metric, which penalizes incorrect estimates for nearby objects more severely than for distant objects. Figure \ref{fig:usvdist_iou_over_dist} shows the mean IoU for increasing distances to the detected objects. As can be seen, the model that received the first place almost always scores highest among its competitors throughout the entire interval. 

Both approaches are based on a modified YOLOv7 architecture capable of predicting distance estimates. The \textit{Data Enhance} method (Section \ref{usv-dist:data_enhance}) extends this model with an additional object detection head, designed specifically for identifying small objects within the image. This enhancement includes the introduction of a small anchor box, tailored for detecting small, distant buoys composed of only a few pixels. Additionally, a BiFormer \cite{liu2024} is incorporated into the architecture, enabling the model to attend to both local and global feature representations, thereby enhancing object detection performance. This improvement is particularly evident in the model’s increased sensitivity to smaller objects, enabled by the attention mechanism. Furthermore, SimAM \cite{yang2021}, a parameter-free attention mechanism that identifies and prioritizes critical neurons, is integrated into the network to boost object detection performance without increasing model complexity. A diverse range of data augmentation techniques, including translation, rotation, flipping, scaling, and mosaic, are applied to the dataset. Among these, data augmentation and the integration of the additional detection head have the most significant impact on improving the combined quality measure.\\
The second-place approach \textit{YOLOv7 Depth-Widened} (Section \ref{usv-dist:depth_widened}) modifies the model architecture by increasing the number of feature channels of each layer, thereby extending the total parameter count to 49 million parameters. Furthermore, the distance loss function is adapted by doubling the loss for objects that are closer than 441 meters, to improve the relative distance error, where nearby objects are penalized more strictly. As with the winning approach a variety of augmentation techniques are employed, specifically horizontal flipping, HSV-augmentation and using a scale parameter to shrink or enlarge the training images. The technique that leads to the most significant improvement of the quality measure for the second place submission is to use data augmentation, in particular setting the scale parameter to improve the performance for distant, barely visible channel markers.

\subsubsection{Discussion and Challenge Winners}
The winners of the USV Supervised Object Distance Estimation challenge are:
\begin{description}
    \item[1\textsuperscript{st} place:] Nanjing University of Science and Technology with \textit{Data Enhance}
    \item[2\textsuperscript{nd} place:] FER LABUST with \textit{YOLOv7 Depth-Widened}
\end{description}
The analysis demonstrates that both methods significantly outperform the baseline model in terms of object detection and distance estimation. Although most architectural modifications were primarily aimed at enhancing object detection performance, the results indicate that these adaptations can also contribute to improvements in distance estimation accuracy.

\subsection{USV-based Obstacle Segmentation Challenge}
\label{sec:usvobstaclesegmentation}

The methods participating in the USV-based Obstacle Segmentation Challenge were required to predict the scene segmentation (into obstacles, water and sky) for a given input image.
The submitted methods have been evaluated on the recently released LaRS benchmark~\cite{Zust2023LaRS}. In addition to the publicly available training set, the authors were also allowed to use additional datasets (upon declaration) for training their methods.

\subsubsection{Evaluation Protocol}
\label{usv-seg:evaluation}

To evaluate segmentation predictions, we employ the LaRS~\cite{Zust2023LaRS} semantic segmentation evaluation protocol. Segmentation methods provide per-pixel labels of semantic components (water, sky and obstacles). However, traditional approaches for segmentation evaluation (\eg mIoU) do not consider the aspects of predictions that are relevant for USV navigation. Instead, the LaRS protocol evaluates the predicted segmentations with respect to the downstream tasks of navigation and obstacle avoidance and focuses on the detection of obstacles.

\begin{table*}
\caption{Performance of the top three submitted segmentation methods and last year's top three best performing methods (denoted in gray) on the LaRS test set. Performance is reported in terms of water-edge accuracy ($\mu$), precision (Pr), recall (Re), F1 score, segmentation mIoU, and overall quality (Q = mIoU $\times$ F1).}
\label{tab:usvseg-results}
\centering
\begin{tabular}{clccccccc}
\toprule
place & method & architecture & Q $\downarrow$ & $\mu$ & Pr & Re & F1 & mIoU \\
\midrule
\mfirst{} & WaterFormer [\S\ref{usv-seg:WaterFormer}] & Mask2Former & \bm1{83.2} & \bm2{80.5} & \bm2{81.6} & \bm1{88.0} & \bm1{84.7} & \bm1{98.3} \\
\msecond{} & WSFormer [\S\ref{usv-seg:WSFormer}] & Mask2Former & \bm2{79.8} & \bm1{80.6} & \bm1{82.9} & 79.8 & \bm2{81.3} & \bm2{98.1} \\
\mthird{} & Advanced K-Net [\S\ref{usv-seg:Advanced_KNet}] & K-Net & \bm3{79.0} & \bm3{79.7} & 77.6 & \bm2{84.2} & \bm3{80.8} & \bm3{97.8} \\
\midrule
\oldfirst{} & \color{gray}SWIM & Mask2Former & 78.1 & \bm3{79.7} & 76.9 & \bm3{83.0} & 79.9 & \bm3{97.8} \\
\oldsecond{} & \color{gray}TransMari & Mask2Former & 77.8 & 79.6 & 78.5 & 82.0 & 80.2 & 97.1 \\
\oldthird{} & \color{gray}Mari-Mask2Former & Mask2Former & 75.7 & 78.4 & \bm3{79.7} & 75.1 & 77.3 & \bm3{97.8} \\
\bottomrule
\end{tabular}
\end{table*}

\begin{figure*}[t]
\centering
   \includegraphics[width=\linewidth]{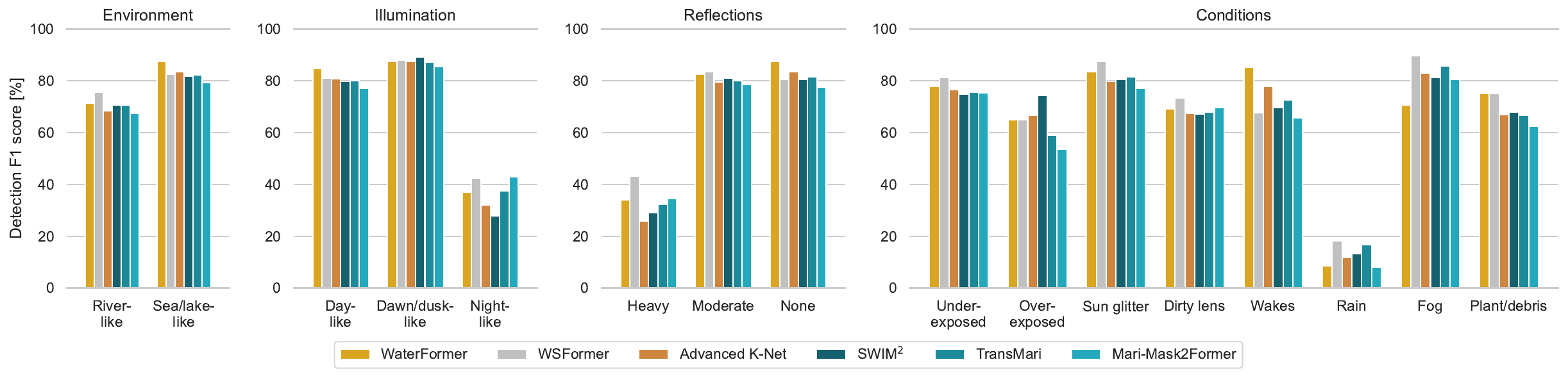}
\caption{Performance of top segmentation methods with respect to different scene attributes.}
\label{fig:usvseg-cat}
\end{figure*}

The detection of static obstacles (\eg shoreline) is measured by the water-edge accuracy ($\mu$), which evaluates the segmentation accuracy around the boundary between the water and static obstacles. On the other hand, the detection of dynamic obstacles (\eg boats, buoys, swimmers) is evaluated by counting true-positive (TP), false-positive (FP) and false-negative (FN) detections, summarized by the F1 score. A ground-truth obstacle is counted as a TP if the intersection with the predicted obstacle segmentation is sufficient, otherwise it is counted as a FN. FPs are counted as the number of segmentation blobs (after connected components) on areas annotated as water in ground truth. For further details, please see \cite{Zust2023LaRS}.

The detection F1 score is a great indicator of the quality of the method predictions, as all obstacles have equal importance in the final score, regardless of their size. However, in the trivial case of predicting everything as an obstacle, all ground-truth obstacles will be counted as TP and there will be only one FP blob per image (albeit very large), which leads to a very large F1 score. On the other hand, mIoU measures the overall segmentation quality on a per-pixel level, but does not reflect the detection of smaller obstacles very well. We thus combine the two measures into a single quality measure (Q = F1 $\cdot$ mIoU) and use it as the primary performance measure in this challenge.

\subsubsection{Submissions, Analysis and Trends}


\begin{figure}[t]
\centering
   \includegraphics[width=\linewidth]{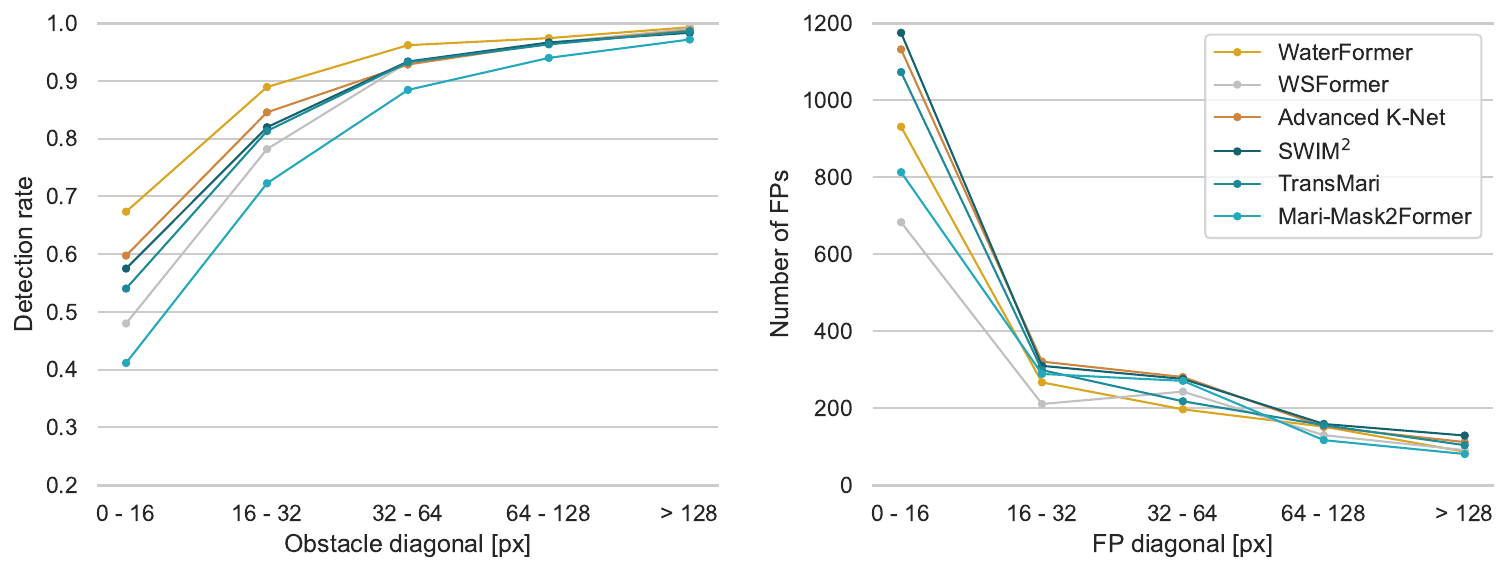}
\caption{Obstacle detection rate and number of FPs across different sizes of obstacles for top performing methods.}
\label{fig:usvseg-sizes}
\end{figure}

\begin{figure*}
    \centering
        \includegraphics[width=\textwidth]{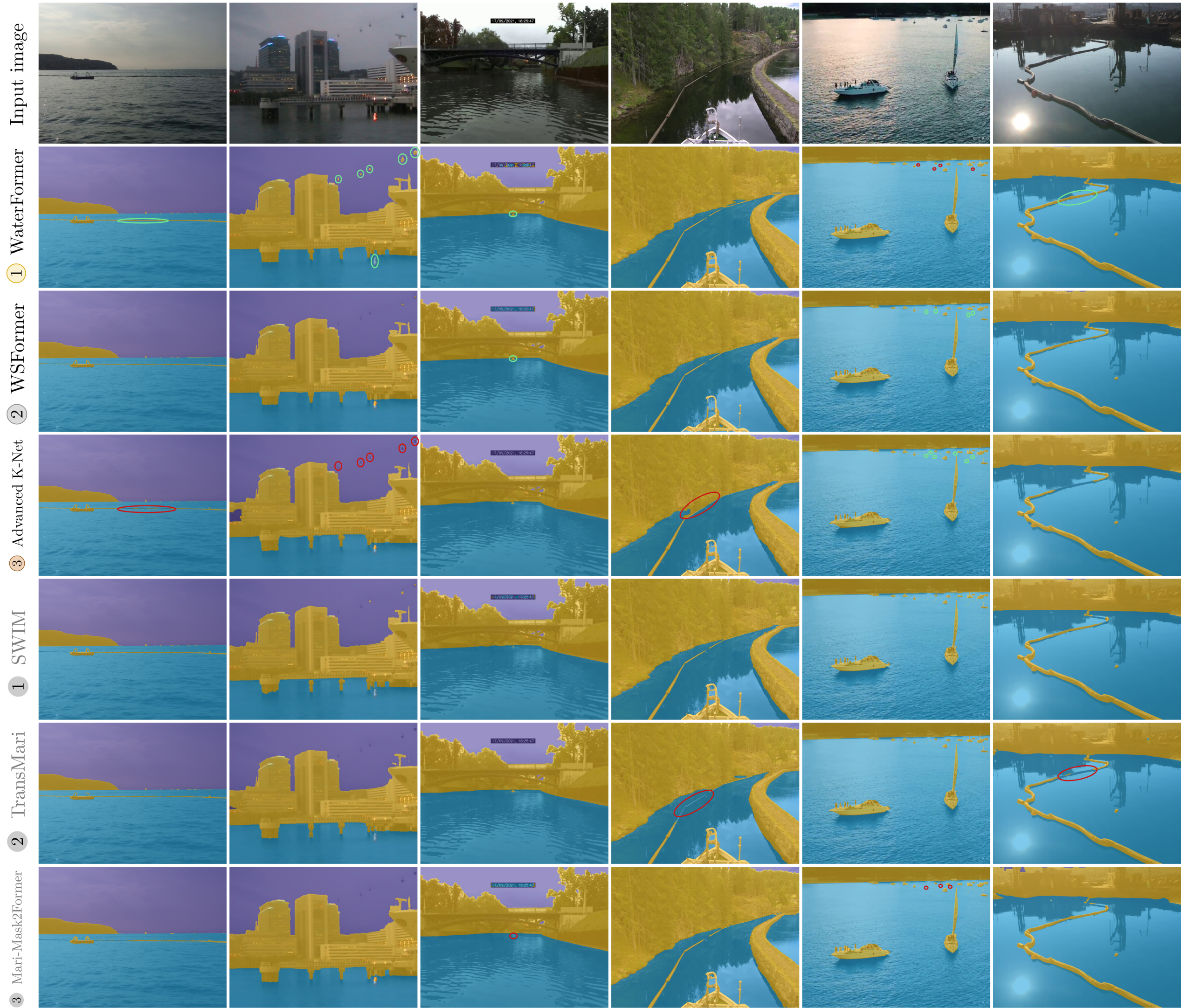}
    \caption{Qualitative comparison of methods for USV-based Obstacle Segmentation. Prominent errors and examples of good behaviour are highlighted in red and cyan, respectively.}
\label{fig:usvseg-qualitative}
\end{figure*}


The USV-based obstacle segmentation challenge received 59 submissions from 16 teams. Only the best performing method from each team is included in the final challenge results. The results of the submitted methods are available on the public leaderboards of the challenge on the MaCVi website~\cite{MaCViWebsite}. Table~\ref{tab:usvseg-results} presents the top three performing methods participating in the challenge, along with the top three performing methods from the previous MaCVi challenge. This section will focus on analyzing the three best-performing submissions, which were the only ones that improved the results of the top performing approach from the previous MaCVi challenge. We refer to teams and their methods by their overall ranking in Table~\ref{tab:usvseg-results} with the notation \rankn{$n$}, where $n$ is the place of the approach. Short descriptions of the top three methods are available in the Appendix~\ref{sec:reports/usvseg}.

This year, three teams improved over the previous year's best performing method, with methods \rankn{1} WaterFormer, \rankn{2} WSFormer, and \rankn{3} Advanced K-Net. The top performing method improved the performance over the last year's winner by 5.1\% Q. Of all the methods in Table~\ref{tab:usvseg-results}, all but one are based on Mask2Former architecture~\cite{cheng2021mask2former}, the exception being Advanced K-Net. WaterFormer included an extra adapter between the layers of DINOv2 backbone layers, and only used LaRS data for training. Additionally, the authors trained the model to predict 4 classes instead of 3 (the obstacle class was split into static and dynamic obstacles, as per the panoptic segmentation labels). The predictions were merged for the evaluation, which was likely the cause of its high performance. The second-best performing method WSFormer also employed Mask2Former architecture, but used the SWIN-L bakcbone. The method was primarily trained on WaterScenes dataset~\cite{yao2024waterscenes}, then fine-tuned on LaRS. Test Time Augmentation was used for producing the final predictions. Advanced K-Net was only trained on LaRS data, using K-Net and a SWIN-L backbone. Lion~\cite{chen2024symbolic} was used instead of Adam for optimizing the model parameters.

\textbf{Detection by scene attributes:}
Figure~\ref{fig:usvseg-cat} 
shows the performance with respect to scene attributes, including environment type, illumination, amount of reflections and scene conditions. Overall, the top three methods outperform the previous challenge winners in most categories, particularly in challenging scenarios such as heavy reflections, foggy scenes and plants or debris presence in water. It should be noted that the second best performing method, WSFormer performs best for many specific attributes, such as night-line illumination, heavy reflections, sun glitter, rain, and fog.

\textbf{Performance by obstacle size:} 
Figure~\ref{fig:usvseg-sizes}
compares the detection performance of the
top three methods and the previous challenge winners across different obstacle sizes. The largest differences between methods are revealed on small obstacles, with the winning method WaterFormer outperforming all others by a quite large margin for very small obstacles.
We can also observe that the second best performing method WSFormer produces the fewest false positives for small obstacles, however that comes at a price of detecting fewer obstacles overall.


\begin{table*}[ht]
\centering
\caption{Results for the USV-based Embedded Obstacle Segmentation Challenge. Methods faster than $\ge 30$ FPS on the target device OAK4 are considered (denoted in teal) when determining the final rankings. For evaluation comparison, we include the winning method from the non-embedded challenge from Section \ref{sec:usvobstaclesegmentation} at the top and last year's~\cite{Kiefer_2024_WACV} winning methods as a baseline, but do not consider them as part of the challenge (denoted in gray). Only the best valid submission from each team is reported. The best results from considered methods are denoted in bold.}
\label{tab:usveseg-overview}
\begin{tabular}{lcrrccccccc}
\toprule
Place & Institution & Method & Section & FPS & $\textbf{Q}\downarrow$ & $\mu$ & Pr & Re & F1 & mIoU \\
\midrule
           & \color{gray} \textit{GIST AI Lab} & \color{gray} \textit{WaterFormer}              & \ref{usv-seg:WaterFormer}             & \color{red} 1.7 (RTX 4090)                            &  83.2 & 80.5	& 81.6	& 88.0	& 84.7	& 98.3 \\ \midrule
\mfirst{}  & DLMU             & RSOS-Net          	   & \ref{usv-eseg:rsos-net}            & \color{teal} 85.1 &  \textbf{64.2} & \textbf{72.5}	& \textbf{66.4}	& 67.9	& \textbf{67.1}	& 95.7 \\
\msecond{}  & ACVLab             & EFFNet          	   & \ref{usv-eseg:effnet}            & \color{teal} 56.3 &  62.3 & 71.3	& 61.1	& 68.0	& 64.4	& \textbf{96.7} \\
           & \color{gray} EAIC-UIT             & \color{gray} eWaSR-RN50          	   &  \cite{Kiefer_2024_WACV}           & \color{teal} \textbf{103.4} &  55.3 & 68.5	& 48.5	& \textbf{70.5}	& 57.4	& 96.2 \\
           & \color{gray} DLMU                 &  \color{gray} Mari-MobileSegNet  		   &   & \textcolor{teal}{60.3}  &  51.3 & 69.1	& 55.7	& 52.2	& 53.9	& 95.3\\
\mthird{}  & EAIC-UIT             & eWaSR-RN101 p2          	   &              & \color{teal} 50.3 &  41.6 & 71.1	& 34.3	& 60.9	& 43.8	& 94.9 \\
           & FarmLabs             & PIDNet-Large          	   &              & \color{red} 21.2 &  41.0 & 57.6	& 40.2	& 61.5	& 48.6	& 84.3 \\

\bottomrule
\end{tabular}
\end{table*}

\textbf{Qualitative results:} Examples of predicted scene segmentations are presented in Figure~\ref{fig:usvseg-qualitative}. We can observe that all the depicted methods interpret the majority of the scene very well, the main differences appearing in the predictions for very small, thin or ambiguous objects, such as the strong reflection in column 2 or the thin structures in columns 1, 4, and 6. WSFormer and Advanced K-Net methods also perform very well on tiny buoys in column 5.


\subsubsection{Discussion and Challenge Winners}


The overall winners of the USV-based Obstacle Segmentation challenge are:
\begin{description}
    \item[1\textsuperscript{st} place:] GIST AI Lab with WaterFormer,
    \item[2\textsuperscript{nd} place:] Multi-institution team (Yancheng Institute of Technology, Hong Kong University of Science and Technology, University of Liverpool,  Xi’an Jiaotong-Liverpool University) with WSFormer, and
    \item[3\textsuperscript{rd} place:] Independent researcher with Advanced K-Net
\end{description}
All three methods outperformed the previous challenge's winner on the LaRS benchmark and show notable improvement, especially with regard to very small objects, thin structures, and low-light circumstances. However, a large number of tiny objects and visually ambiguous appearances remain open issues. We expect the performance will continue to improve with new datasets and newer methods.


\subsection{USV-based Embedded Obstacle Segmentation}
\label{sec:usvembeddedobstaclesegmentation}

Modern obstacle detection methods often depend on high-performance, energy-intensive hardware, making them unsuitable for small, energy-constrained USVs~\cite{tervsek2023ewasr}. The USV-based Embedded Obstacle Segmentation challenge aims to address this limitation by encouraging development of innovative solutions and optimization of established semantic segmentation architectures which are efficient on embedded hardware. The challenge builds on the success of last year's challenge~\cite{Kiefer_2024_WACV} and is an extension of the USV-based Obstacle Segmentation task outlined in Section~\ref{sec:usvobstaclesegmentation}. Submissions are evaluated and benchmarked on a real-world OAK4 device from Luxonis.

\subsubsection{Evaluation Protocol}



To ensure compatibility with embedded devices, we impose additional constraints that must be adhered to during the development and submission phases like in ~\cite{Kiefer_2024_WACV}. Submissions must follow:
\begin{itemize} 
    \item \emph{Static Graph} -- Neural networks must have a well-defined static graph and fixed shapes, and must be exportable to ONNX format without custom operators. 
    \item \emph{Operation Support} -- ONNX can contain only a specific set of operations constrained by the hardware, details of which we provide before the start of the challenge. 
    \item \emph{Standardized inputs and outputs} -- Models are evaluated directly on the device, so they must accept fixed-size input images ($768 \times 384$) normalized using ImageNet~\cite{deng2009imagenet} mean and standard deviation. Output must be a single-channel 2D segmentation mask containing per-pixel class indices. Only methods with a single image input and a single output are permitted. 
    \item \emph{Performance Requirement} -- Models must achieve a throughput of at least $30$ FPS on the embedded device OAK4.
\end{itemize}

Submitted architectures are quantized to INT8 using the validation set of LaRS~\cite{Zust2023LaRS} and leveraging Qualcomm's SNPE toolkit~\cite{snpe}. The models are then compiled into a binary executable for the target hardware. During inference, the input images are resized, centered, and padded to a $768 \times 384$ resolution with the maintained aspect ratio, and the outputs are resized to match the original image resolution. The final scores are computed using the same evaluation protocol described in \ref{usv-seg:evaluation}. In addition to the LaRS metrics, the average throughput achieved on the embedded device is also reported.

\subsubsection{Submissions, Analysis and Trends}


The embedded obstacle segmentation challenge received $26$ submissions from $4$ different teams. In addition, we consider the winning method from last year's competition. We show the best submission from each team in Table~\ref{tab:usveseg-overview}. For final rankings, we only consider submissions from this year which are faster than $30$ frames per second. We received submission reports for the $2$ winning methods.

\begin{figure}[ht]
\centering
   \includegraphics[width=0.8\linewidth]{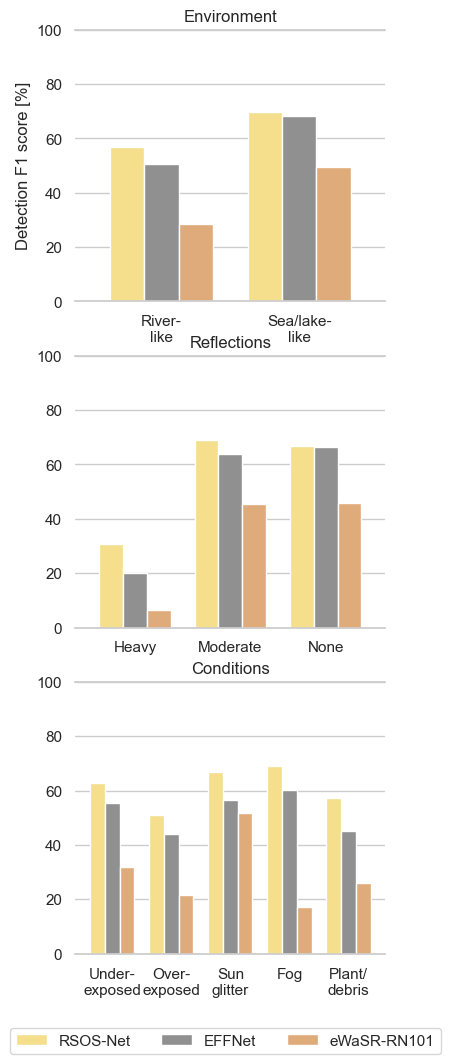}
\caption{Comparison of F1 detection score for the best $3$ submissions from Section~\ref{sec:usvembeddedobstaclesegmentation} under various environmental conditions (environment type, reflections, and conditions).}
\label{fig:usveseg-envs}
\end{figure}


\begin{figure*}[pht]
    \centering
    \vspace{2em}
    \begin{tabular}{>{\centering\arraybackslash}p{0.22\textwidth}>{\centering\arraybackslash}p{0.22\textwidth}>{\centering\arraybackslash}p{0.22\textwidth}>{\centering\arraybackslash}p{0.22\textwidth}}
    (a) Input image & (b) RSOS-Net & (c) EFFNet & (d) eWaSR-RN101 \\
    \end{tabular}
    \includegraphics[width=\textwidth]{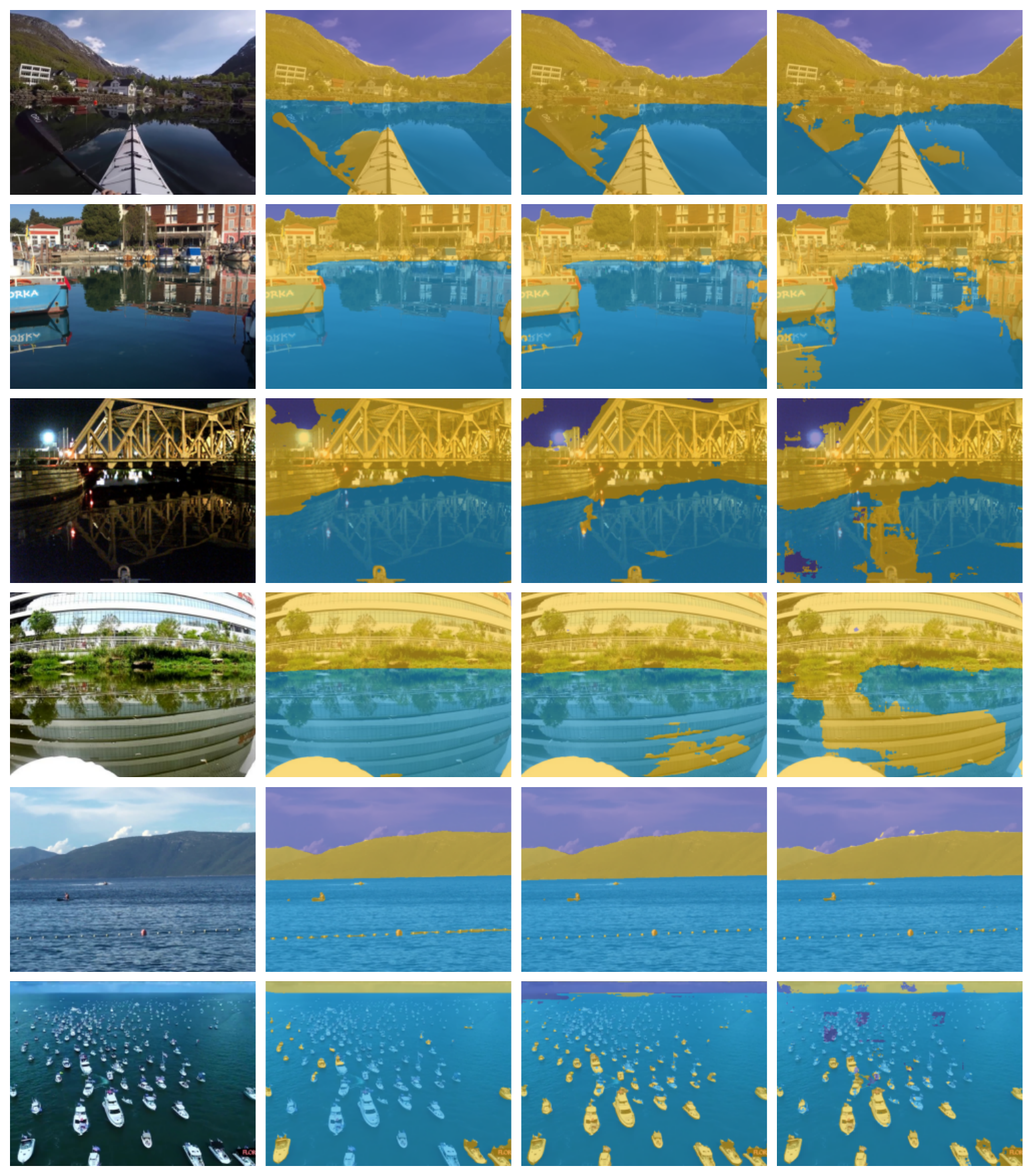}
    \caption{Qualitative comparison of the best $3$ submissions for USV-based embedded obstacle segmentation from Section~\ref{sec:usvembeddedobstaclesegmentation}.}
\label{fig:usveseg-qualitative}
\end{figure*}

Both RSOS-Net (\ref{usv-eseg:rsos-net}) and EFFNet (\ref{usv-eseg:effnet}) prioritize lightweight architectures and efficient real-time inference leveraging ResNet~\cite{he2016deep} backbones. RSOS-Net excels at balancing speed and accuracy, while EFFNet explores ensemble techniques to enhance prediction robustness.

RSOS-Net (Section~\ref{usv-eseg:rsos-net}) features a ResNet-101 backbone paired with a lightweight decoder employing separable convolutions and an attention-based fusion module. Its Fast Pyramid Pooling Module (FPPM) enhances the receptive field, effectively distinguishing obstacles from surface disturbances. EFFNet (Section~\ref{usv-eseg:effnet}) utilizes an ensemble of two models -- ResNet-50 and ResNet-101 backbones combined with DeepLabv3+~\cite{chen2018deeplab} decoders. Prolonged training improved performance, while additional augmentations or heavier backbones did not. The method emphasizes multi-scale feature extraction and stability after quantization. No additional data is used in both methods and they are trained only on the LaRS~\cite{Zust2023LaRS} dataset.

Both winning methods significantly outperformed last year's winner and baseline method eWaSR-RN50 ($+8.9$ Q and $+7.0$ Q for RSOS-Net and EFFNet, respectively). At a small cost in recall ($-2.6\%$ and $-2.5\%$), precision was significantly better ($+17.5\%$ and $12.6\%$). Both methods were slower ($85.1$ FPS and $56.3$ FPS for RSOS-Net and EFFNet, respectively) than baseline ($103.4$ FPS eWaSR-RN50) but still achieved high throughput and are suitable for embedded device deployment. We provide full quantitative results in Table~\ref{tab:usveseg-overview}.

Additionally, we compare the three winning methods under various environmental conditions. For readability, we only include the most interesting results in Figure~\ref{fig:usveseg-envs}. As expected, both RSOS-Net and EFFNet significantly outperform other methods in various scenarios. However, RSOS-Net and EFFNet mostly differ in a few key areas. RSOS-Net is considerably better in the presence of heavy reflections. It also outperforms EFFNet under challenging conditions, such as over or under-exposure, glitter, fog, and plant debris. The latter could be a reason for the slightly better performance of RSOS-Net in river-like scenes.

We further qualitatively compare the best $3$ submissions in Figure~\ref{fig:usveseg-qualitative}. RSOS-Net indeed outperforms EFFNet under conditions discussed in the previous paragraph. We can see it performs better in the presence of reflections and during night. While all methods are capable of detecting small objects like buoys, they are still limited by the fixed input shape requirement and at times struggle to detect small objects when multiple are present in the scene.

\subsubsection{Discussion and Challenge Winners}

The overall winners of the USV-based Embedded Obstacle Segmentation are:
\begin{description}
    \item[1\textsuperscript{nd} place:] Dalian Maritime University (DLMU) with RSOS-Net.
    \item[2\textsuperscript{nd} place:] Advanced Computer Vision Lab (ACVLab) with EFFNet, and 
    \item[3\textsuperscript{rd} place:] University of Information Technology, VNU-HCM (EAIC-UIT) with eWaSR-RN101 p2.
\end{description}


The $2$ best-performing submissions significantly outperformed winning methods from last year, while maintaining above real-time throughput. Submitted methods adapt better to challenging scenarios such as glitter, reflections, and lightning. However, due to fixed input shape requirements, they still perform worse in the presence of many small obstacles, especially compared to regular segmentation track. We leave this open for future competitions.


\subsection{USV-based Panoptic Segmentation Challenge}
\label{sec:usv-panoptic-challenge}

\begin{table*}[ht]
\centering
\caption{Results for the USV-based Panoptic Segmentation Challenge measured in overall panoptic quallity (PQ) and separate for thing and stuff classes. Competing methods are compared to a baseline Mask2Former model~\cite{cheng2021mask2former} and the top-performing PanSR method~\cite{Zust2024PanSR}. *The throughput of methods, measured in frames-per-second (FPS), is self-reported by the authors.}
\label{tab:usv-panoptic-overview}
\begin{tabular}{ccrrcccccc}
\toprule
Place & Team & Method & Section & Hardware & FPS* & PQ & PQ$_\text{th}$ & PQ$_\text{st}$ \\
\midrule
 $\textcolor{pink}{\bigstar}$ & MaCVi Team & PanSR (Swin-L)             &     \S\ref{usv-pan:baselines}         & RTX A4500 & 0.7                           &  \bm1{57.3} &	\bm1{43.0}  &	\bm2{95.4} \\
\midrule
\mfirst{}  & Fraunhofer IOSB         & MaskDINO         &  \S\ref{usv-pan:first}          & V100 & 1.0 &  \bm2{53.9} &	\bm2{39.4} 	& 92.3 \\
\msecond{}  & GIST AI Lab         & PanopticWaterFormer         &  \S\ref{usv-pan:second}           & H100 & 1.1 &  \bm3{53.0} 	& \bm3{37.3} 	& 94.7 \\
\mthird{}  & UL~1      &  FC-CLIP-LaRS   &  \S\ref{usv-pan:third}           & A100 & 10 &  49.7 	& 32.4 	& \bm1{95.6} \\
    4th     & WaterScenes      &  WaterScenes   &  -          & RTX 4090 & 10 &  48.6 	& 31.2 	& \bm3{95.1} \\
    5th     & YCIT      &  SeeDroneSee   &  -          & H100 & 4.8 &  47.7 	& 30.0 	& 95.0 \\
    6th     & XJTLU      &  Mask   &  -          & RTX 3090 & 1.7 &   47.4 	& 29.5  &	94.9 \\
    7th     & UL~2  &  PSAM   &  -          & RTX 2080 Ti & 0.6 &   46.2 	& 28.9 	& 92.1 	 \\
\midrule
\color{gray} - & \color{gray} MaCVi Team  &  \color{gray} Mask2Former (Swin-B)   &  -          & V100 & 4.8 &   41.4  &	22.3  &	92.5 	 \\

\bottomrule
\end{tabular}
\end{table*}

\begin{figure*}[t]
\centering
   \includegraphics[width=\linewidth]{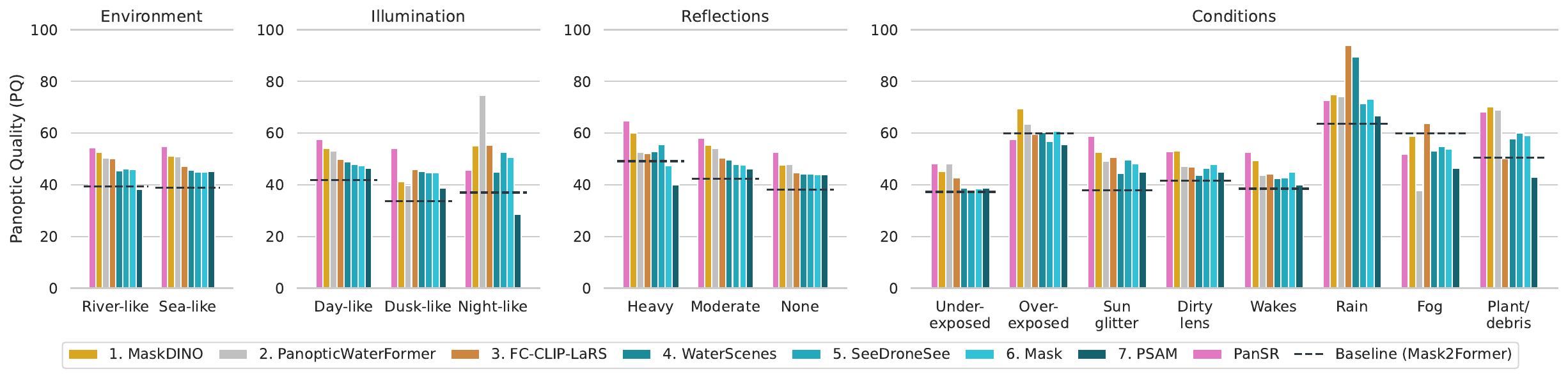}
\caption{USV-based Panoptic Segmentation: Performance of top methods in terms of panoptic quality (PQ) w.r.t. different scene attributes. The performance of the baseline method (Mask2Former~\cite{cheng2021mask2former} w/ Swin-B) is marked as a dotted line for reference.}
\label{fig:usvpan-cat}
\end{figure*}

\begin{figure}[t]
\centering
   \includegraphics[width=\linewidth]{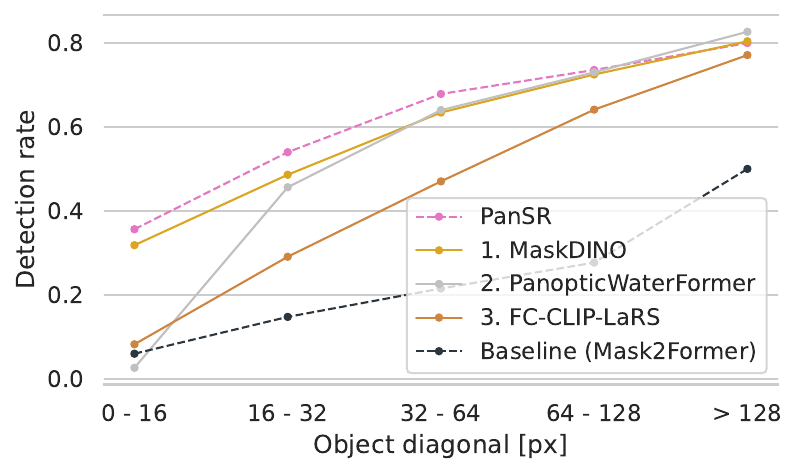}
\caption{USV-based Panoptic Segmentation: Dynamic obstacle detection rate of top methods with respect to ground-truth obstacle size.}
\label{fig:usvpan-size}
\end{figure}

This year MaCVi introduced a new panoptic sub-challenge based on the recent LaRS dataset~\cite{Zust2023LaRS}. In contrast to our segmentation challenges, the panoptic challenge calls for a more fine-grained parsing of USV scenes, including segmentation and classification of individual obstacle instances. This formulation encapsulates the requirements of scene parsing for USV navigation in a more principled way, paving the road for downstream tasks such as tracking individual obstacles, trajectory prediction and obstacle avoidance.

The task of the USV-based Panoptic Segmentation Challenge was to develop a panoptic segmentation method that parses the input USV scene into several background (\textit{i.e.} stuff) and foreground (\textit{i.e.} things) classes and outputs a set of instance masks and their class labels. Stuff classes include water, sky and static obstacles, and thing classes include 8 different types of dynamic obstacles -- boat/ship, buoy, row boat, swimmer, animal, paddle board, float and other. In addition to training on the LaRS training set, which uses the aforementioned classes, participants were allowed to use other datasets for (pre-)training if disclosed.

\subsubsection{Evaluation protocol}

The panoptic performance of submitted methods is evaluated using standard panoptic performance evaluation measures~\cite{Zust2023LaRS,Kirillov2019Panoptic}: segmentation quality (SQ), recognition quality (RQ) and the combined panoptic quality (PQ):
%
\begin{equation}
     \text{PQ} = \underbrace{\frac{\sum_{(p,g) \in \text{TP}} \text{IoU}(p, g)}{|\text{TP}|}}_\text{segmentation quality (SQ)} \times \underbrace{\frac{|\text{TP}|}{|\text{TP}| + \frac{1}{2}|\text{FP}| + \frac{1}{2}|\text{FN}|}}_\text{recognition quality (RQ)}.
\end{equation}
%
To determine the ranking of the submitted methods, overall panoptic quality (PQ) was considered. For additional insights, individual metrics are also reported separately for \textit{thing} and \textit{stuff} classes indicated by superscripts $(\cdot)^\text{Th}$ and $(\cdot)^\text{St}$. 
 

Following~\cite{Zust2023LaRS}, additional dynamic obstacle detections inside static obstacle regions are not considered as false positives. Finally, every participant was required to submit information about the speed of their method including the hardware used for benchmarking and the measured speed in frames-per-second (FPS).

\subsubsection{Submissions, Analysis and Trends}

\begin{figure*}[t]
\centering
   \includegraphics[width=\linewidth]{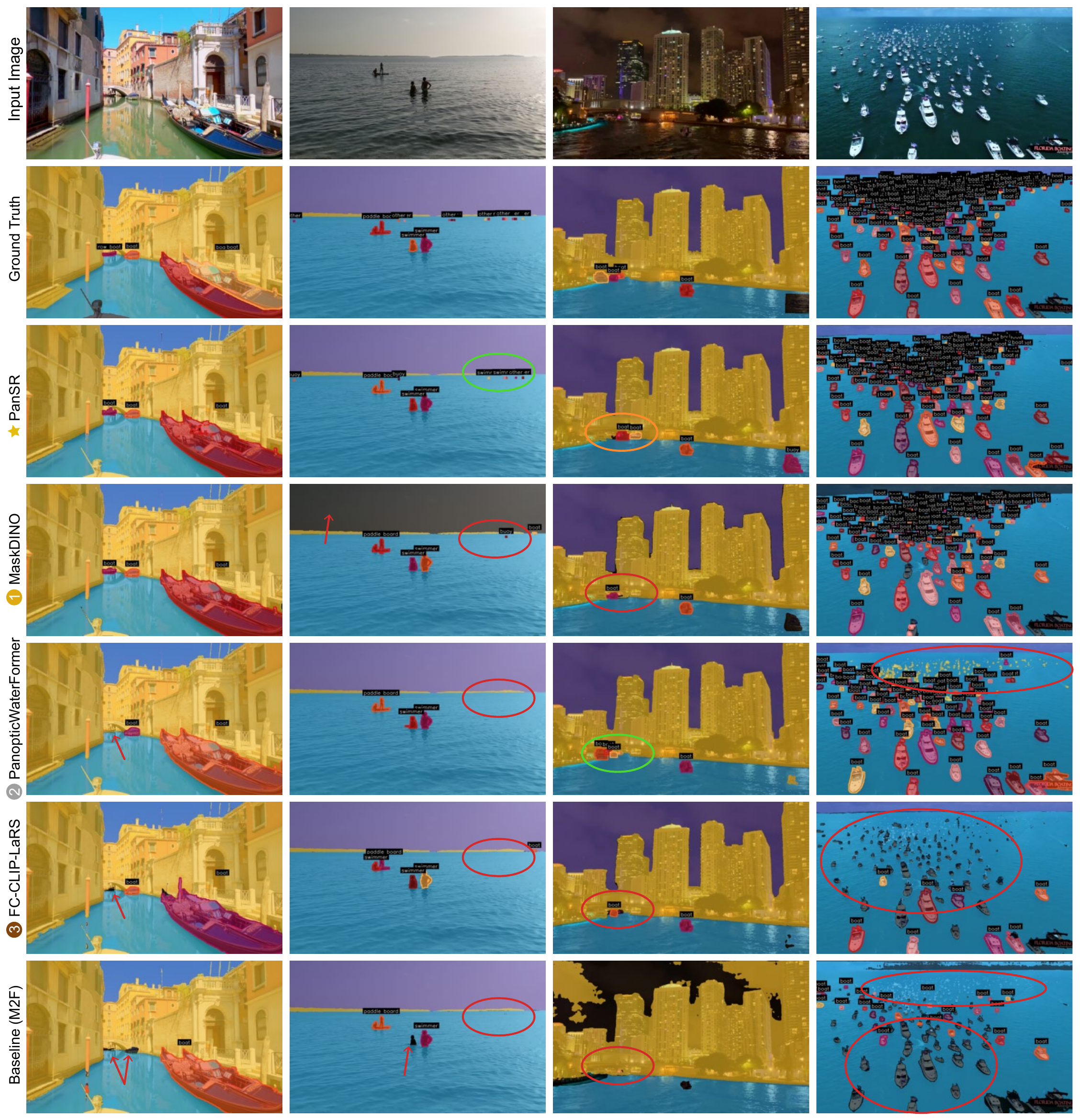}
\caption{USV-based Panoptic Segmentation: Qualitative comparison of top 3 ranking methods.}
\label{fig:usvpan-qual}
\end{figure*}

The USV-based Panoptic Segmentation Challenge received 21 submissions in total from 7 different teams. We only analyze the best-performing method of each team as per the rules of the challenge. Results of the remaining submitted methods are available on the public leaderboards of the challenge on the MaCVi website~\cite{MaCViWebsite}. In Table~\ref{tab:usv-panoptic-overview} we report the overall results and final standings of the best methods submitted by each team. The MaCVi team also contributed a baseline Mask2Former (Swin-B) method~\cite{cheng2021mask2former} for comparison. Additionally, a recent PanSR method~\cite{Zust2024PanSR} was submitted to the challenge by the MaCVi team, but is not considered for the winning positions of the challenge. In this section we analyze all the contributed methods, with a special focus on the top three approaches. Short descriptions of the top three methods and baseline approaches are also available in the Appendix~\ref{usv-pan:submissions}.

We are happy to report that all the teams contributed strong methods and improve over the baseline Mask2Former method by a significant margin. Specifically, the first-placing MaskDINO method~\cite{Li2022Mask} achieves a PQ of 53.9, outperforming the baseline by +12.5 \% PQ. While not considered for winning the challenge, the top-performing PanSR method boosts performance by an additional +3.4 \% PQ.

Interestingly, the top three methods take a vastly different approach to the problem, but reach similar performance. The \rankn{3} method utilizes an open-vocabulary CLIP-based method~\cite{yu23neurips}, the \rankn{2} method combines a top-performing segmentation network from the Obstacle Segmentation Challenge (Section~\ref{sec:usvobstaclesegmentation}) with an instance segmentation network~\cite{cai2019cascade} for addressing thing classes, and the \rankn{1} method utilizes a strong detection-optimized panoptic network~\cite{Li2022Mask}, while also experimenting with an open vocabulary approach. None of analysed methods utilized additional data during training other than standard pre-training datasets (COCO~\cite{coco}). In the following, we analyze the top-performing methods of each of the teams in more detail.

\textbf{Detection by scene attributes:} 
In Figure~\ref{fig:usvpan-cat} 
we compare the performance of methods with respect to scene attributes, including environment type, illumination, amount of reflections and scene conditions. Overall, all submited methods outperform the baseline (Mask2Former~\cite{cheng2021mask2former}) in most categories. The top performing methods also demonstrate great stability across different scenarios. On several exceedingly rare scene attributes ($<5\%$) such as night-like illumination, rain and fog, performance of methods is less consistent with overall trends. For example, on night-time scenes the \rankn{2} method PanopticWaterFormer achieves the best performance, also visible in the qualitative examples (Figure~\ref{fig:usvpan-qual}).

\textbf{Performance by obstacle size:} 
Figure~\ref{fig:usvpan-size} 
compares the detection performance of the
top three methods with the Mask2Former baseline~\cite{cheng2021mask2former} and the top-performing PanSR method~\cite{Zust2024PanSR} across different obstacle sizes. 
As expected, the largest differences between methods are revealed on small obstacles. Specifically, PanSR and the \rankn{1} MaskDINO method offer a major boost in the detection rate on the smallest obstacles compared to other methods. 

\textbf{Qualitative results:} Examples of panoptic method predictions for top methods are shown in Figure~\ref{fig:usvpan-qual}. Compared to the baseline (Mask2Former) all methods improve the segmentation quality and detect a larger percentage of dynamic obstacles. The largest differences between methods are observed on small obstacles (column 2 and 4) and in crowded scenes (column 4), where \rankn{1} MaskDINO and especially PanSR perform significantly better.

\subsubsection{Discussion and Challenge Winners}

The overall winners of the USV-based Panoptic Segmentation challenge are as follows:
\begin{description}
    \item[1\textsuperscript{st} place:]{Fraunhofer IOSB with MaskDINO,}
    \item[2\textsuperscript{nd} place:]{GIST AI Lab with PanopticWaterFormer,}
    \item[3\textsuperscript{rd} place:]{University of Ljubljana 1 with FC-CLIP-LaRS}
\end{description}

Although using very different approaches, all three methods outperformed the baseline method on the LaRS benchmark and show notable improvement, especially with regard to segmentation quality and detection of small objects. However, these methods still struggle in the case of crowded scenes, very small objects and rare cases or conditions. The recent PanSR method~\cite{Zust2024PanSR} already demonstrates that architectural and training choices can make a large impact on the detection performance on crowded scenes and small objects. Furthermore, some teams experimented with an open-vocabulary approach, which shows potential for addressing rare cases by harnessing language-grounded knowledge from large-scale data.

\section{MarineVision Restoration Challenge}
\label{sec:uir-challenge}
Underwater image restoration \cite{Desai_2021_CVPR, Desai_2022_CVPR, Desai_2023_WACV, Desai_2023_ICCV} is the process of enhancing the visual quality of images captured in aquatic environments by eliminating degradations caused by light absorption and scattering. These degradations reduce contrast and introduce color casts, making object detection and localization of underwater species a challenging task. Restoring lost visibility and improving color balance aid in the detection and localization of underwater species, facilitating tasks such as biodiversity monitoring, marine conservation, and underwater robotics. The MarineVision Restoration Challenge (MVRC) focuses on developing robust image restoration methods to enhance the detection and localization of underwater species. In the first edition of the MVRC challenge the participants are provided with the RSUIGM dataset \cite{rsuigm}. The dataset consists of 30 ground truth scenes and 200 corresponding synthetically generated degraded observations for each scene.
\begin{figure*}
    \centering
    \includegraphics[width=1\linewidth]{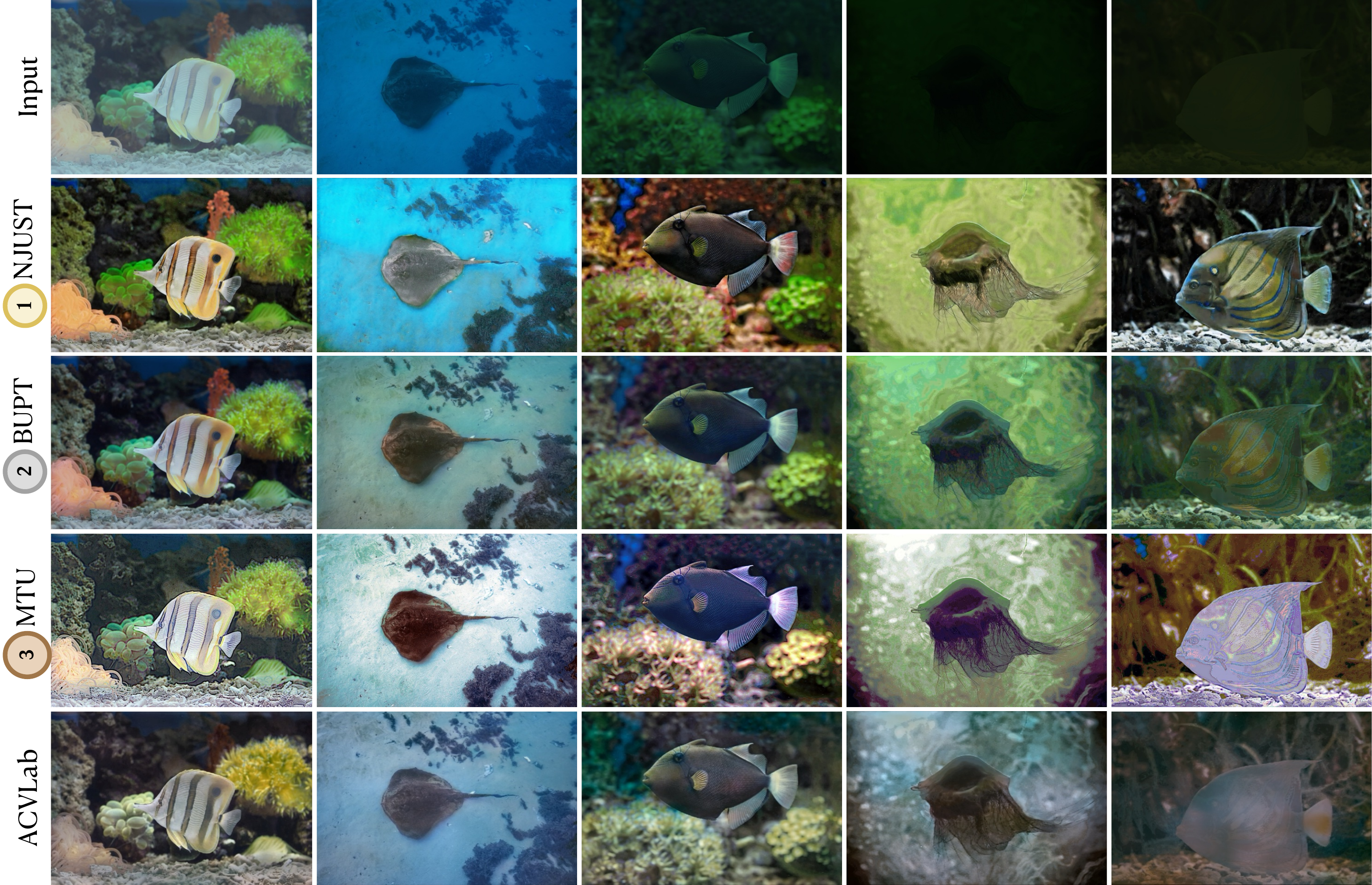}
    \caption{Qualitative comparison of underwater image restoration proposed by top performing teams in MarineVision Restoration Challenge.}
    \label{fig:mvrc_qual_results}
\end{figure*}
The challenge is divided into two steps: in the first step, participants must develop a robust algorithm for restoring the degraded images; in the second step, participants must perform the detection and localization of the underwater species considering the restored images from the first step. Participants are permitted to utilize other publicly available datasets to enhance underwater image restoration.
\subsection{Evaluation Protocol}
The evaluation is conducted in two phases. In the first phase,  we consider perceptual quality metrics (PQM)-- Underwater Image Quality Metric (UIQM), Underwater Color Image Quality Evaluation (UCIQE), Colorfulness Contrast Fog density index (CCF) to assess the performance of restoration methods. In the second phase, we consider Mean IoU (mIoU), F1 Score, and mean Average Precision (mAP) performance metrics to evaluate the effectiveness of the detection (OD) and localization methods. Further, we compute the overall score Q as shown below: 
\begin{align}
    Q &= 0.8 \left( \frac{\text{CCF} + \text{UIQM} + \text{UCIQE}}{3} \right) \nonumber \\
      &\quad + 0.2 \left( \frac{\text{F1 score} + \text{mAP} + \text{mIoU}}{3} \right)
\end{align}
\textbf{Note}: The CCF metric does not have a defined range of values. Therefore, we normalize CCF values between 0 and 1 for fairness and ease of evaluation. We emphasize more on perceptual enhancement of the images and showcase marine species detection as a downstream application of underwater image restoration, hence the weightage (80\%) for perceptual metrics and (20\%) for object detection evaluation metrics.  
\subsection{Submissions, Analysis and Trends}
In the first edition of the challenge, we received over 40 submissions from 8 different teams. The top performing teams are reported in Table \ref{tab:mvrc_results}. The top 4 teams were invited to submit the description of their methods, which are provided in sections \ref{uir:first}, \ref{uir:second}, \ref{uir:third}, \ref{uir:fourth} respectively.  
The submissions show the authors refer to a wide variety of models and approaches for underwater image restoration and enhancement. Team (\#1) NJUST-KMG adopt a two-stage pipeline for underwater image enhancement. The authors first restore the underwater images using HVI-CIDNet and to further remove the hazy effect in the images, the authors employ dehazing methods CasDyF-Net and SCANet. Followed by restoration and enhancement, the authors perform marine species detection using pre-trained version of DETR object detection model. In contrast other teams (\#2) BUPT, (\#3) MTU, and (\#4) ACVLab use end to end approaches for underwater image restoration and enhancement followed by object detection. Team (\#2) BUPT, build upon the existing NAFNet with a frequency separation and fusion block aimed at processing high and low frequency details separately. While teams (\#1) and (\#2) adopt deep learning approaches, team (\#3) adopt statistical method, Texture Enhancement Model Based on Blurriness and Color Fusion (TEBCF) to restore degraded underwater images. TEBCF is a non-data driven method, requires no pre-training and can be directly used for inferencing. Followed by restoration, the authors use pre-trained YOLO-NAS model for detecting marine species in the restored underwater images. The qualitative comparison of under image restoration methods proposed by top performing teams in shown in Figure \ref{fig:mvrc_qual_results}.

\subsection{Discussion and Challenge Winners}

\begin{table*}[!h]
    \centering
    \resizebox{\linewidth}{!}{
    \begin{tabular}{c l r c c c c c c c c c c c}
        \hline
        \textbf{Place} & \textbf{Team} & \textbf{Method} & \textbf{Section} & \textbf{Hardware} & \textbf{UIQM} & \textbf{UCIQE} & \textbf{CCF} & \textbf{mIoU} & \textbf{F1} & \textbf{mAP} & \textbf{PQM} & \textbf{OD} & \textbf{Final Score} \\ \hline
        \protect\mfirst{} & NJUST-KMG & NJUST-KMG & \S\ref{uir:first} & RTX 3090 & 0.70 & 0.57 & 0.08 & 0.69 & 0.13 & 0.13 & 0.45 & 0.31 & 0.42 \\ 
        \protect\msecond{} & BPUT MCPRL & BUPT MCPRL & \S\ref{uir:second} & RTX 4090 & 0.37 & 0.53 & 0.02 & 0.64 & 0.16 & 0.16 & 0.31 & 0.32 & 0.31 \\ 
        \protect\mthird{} & MTU & MTU & \S\ref{uir:third} & Tesla V100 & 0.00 & 0.60 & 0.01 & 0.64 & 0.16 & 0.16 & 0.20 & 0.32 & 0.23 \\ 
        4 & ACVLab & Color Transfer Fusion Network & \S\ref{uir:fourth} & RTX 2080 Ti & 0.00 & 0.50 & -0.01 & 0.54 & 0.14 & 0.14 & 0.16 & 0.27 & 0.18 \\ 
        \hline
    \end{tabular}
    }
    \caption{Quantitative comparison of methods proposed by top teams in MarineVision Restoration Challenge. The final score is measured as the weighted average of Perceptual Quality Metrics (PQM) and Object Detection (OD) metrics.}
    \label{tab:mvrc_results}
\end{table*}
The winners of the challenge are as follows: 
\begin{description}
    \item[1\textsuperscript{st} place:]{Nanjing University of Science and Technology}
    \item[2\textsuperscript{nd} place:]{Beijing University of Posts and Telecommunications, Beijing, China with MCPRL}
    \item[3\textsuperscript{rd} place:]{Michigan Technological University, USA}
\end{description}
All the Top Three teams outperformed the set baselines for underwater image restoration task. Team (\#1) demonstrated a two-stage pipeline for restoration--image restoration followed by haze removal, significantly improving the performance in the detection of marine species. In contrast, Team (\#3) showcased a non-learning-based algorithm for underwater image restoration. However, the resulting images consist of sharp edges in all images, indicating the need for
better restoration models.

\section{Conclusion}

In this summary paper, we analyzed the challenges of the 3\textsuperscript{rd} Workshop on Maritime Computer Vision. Specifically, MaCVi 2025 hosted four different surface domain maritime challenges, and one underwater image restoration challenge.

The USV-based Object Distance Estimation challenge focused on detecting buoys in image space and their metric distances in world space. The top two submission significantly outperformed the baseline yielding insights into backbone selection and loss function balance. In future versions, it might be interested to see the more complicated multi-class or open vocabulary distance estimation setups.

USV-based Obstacle Segmentation Challenge addressed semantic scene parsing for USVs with a focus on obstacle detection and saw significant improvements over submissions of the previous iteration of MaCVi. 

USV-based Embedded Segmentation Challenge evaluated method on the same task, but running it on real-world embedded hardware and all its limitations. While still far from the unbounded state-of-the-art, this year's methods substantially reduced the performance gap at impressive, more than real-time speeds, and are quickly catching up to the top-performing methods. 

USV-based Panoptic Segmentation featured a much more difficult panoptic task, which requires methods to detect, segment and classify also individual instances of dynamic obstacles in the scene. All participants significantly outperformed the strong baseline we set at the start of the challenge and made large leaps in tackling major challenges of the task, such as the detection of very small obstacles and robustness to variation in scene conditions.

The First Edition of the MarineVision Restoration Challenge addressed the complex task of underwater perception. The challenge provided synthetic data that largely resembled the real-world optical properties of water pertaining to diverse geographical locations. The task involved restoration of images and the perform marine species detection on the restored images as a downstream application useful in tasks such as marine life health monitoring and conservation, perception for underwater vehicles such as submarines and ships. 
All participants outperformed the set baselines for the underwater image restoration task.

Overall, the challenges presented at MaCVi 2025 featured strong entries across the board, pushing the boundaries of what is possible in maritime vision in several different aspects, while outlining multiple exciting future research directions for the field.

\vspace{.2cm}
\noindent\textbf{Acknowledgments.}
This work was supported by
Conservation, Protection and Use joint call,
Slovenian Research Agency (ARIS) project J2-60054 and programs P2-0214 and P2-0095,
Shield AI for sponsoring prizes, and LOOKOUT for providing testing videos for the USV-based Object Distance Estimation challenge.


\appendix
\addcontentsline{toc}{section}{Submitted Methods}
\section*{Appendix - Submitted Methods}


\section{Supervised Object Distance Estimation}

\subsection{\protect\mfirst{} Data Enhance}
\label{usv-dist:data_enhance}
\emph{Zhiqiang Zhong, Zhe Zhang, Sujie Liu, Xuran Chen, Yang Yang}\\
\emph{Nanjing University of Science and Technology}\\
\texttt{\small{\{zhongzhiqiang2002, zhe.zhang, xuran\_chen, yyang\}}@njust.edu.cn}, \texttt{sruis1109@163.com}\\
\\
This report presents the solution to the Approximate
Supervised Object Distance Estimation challenge at the
3rd Workshop on Maritime Computer Vision. We use the
YOLOv7 base model, enhancing it with an additional detec tion head and the BiFormer attention mechanism to better focus on small objects. Additionally, we increase the sample size by horizontally flipping images and apply a range of data augmentation techniques to improve the model’s performance. We also integrate the SimAM attention mechanism to further enhance detection performance. Ultimately, we achieve a score of 0.2719, securing first place.\\
\textbf{Introduction:} We focus on several key areas,
including multimodal machine learning, open-environment
machine learning, and incremental \cite{yangyang2022, yangyang2023, yangyang2024}. This challenge involves developing algorithms for estimating the distance to buoys from monocular images.\\
\textbf{Method:} This section provides an overview of our approach. Our
method is built upon the YOLOv7 model and primarily includes data augmentations, an additional detection head, the BiFormer attention and the SimAM attention.
\begin{itemize}
    \item \textbf{Data Augmentations:} We increase the sample size by horizontally flipping the images and enhance the model’s performance and robustness using a range of data augmentation techniques, including image rotation, translation, scaling, shearing, perspective transformation, vertical and horizontal flipping, mosaic, mixup, and copy-paste.
    \item \textbf{Additional Detection Head:} YOLOv7 is an anchor-based object detection model, rather than anchor-free. Therefore, we add a small initial anchor box and incorporate an additional detection head to detect small objects in the image.
    \item \textbf{BiFormer:} BiFormer \cite{liu2024} is an innovative attention mechanism that enhances the model’s detection performance by simultaneously capturing both local and global features through a dual-layer routing design. During the feature extraction process, BiFormer computes attention for each local region to capture detailed features. This step improves the model’s sensitivity to small objects and fine details, thereby enhancing detection accuracy. We integrate BiFormer Attention as a plugin into YOLOv7. Specifically, we add BiFormer Attention at the end of the SPPCSPC module and also at the end of the Conv module.
    \item \textbf{SimAM:} SimAM \cite{yang2021} is a parameter-free attention mechanism inspired by neuroscience that assigns 3D attention weights to feature maps in a flexible manner, without increasing model complexity. This enables more efficient and lightweight processing. By using an energy function to measure the linear separability between neurons, SimAM identifies and prioritizes important neurons, thereby enhancing feature extraction. We integrate SimAM into a modified YOLOv7 model to improve object detection performance, leveraging its ability to achieve attention without the need for additional parameters.
\end{itemize}
\textbf{Experiments:} The object distance estimation dataset comprises about 3,000 images of maritime navigational aids, mainly featuring red and green buoy markers. Only the training set is provided, while the test set is withheld to benchmark model performance in the competition.
We train the model using an NVIDIA GeForce RTX 3090, with the number of 400. The experimental results are shown in the table below.
{\scriptsize
\begin{tabularx}{\linewidth}{cc}
    \hline
     Method &  Scores\\
     \hline \hline
     yolov7 &  0.2260\\
     + augmentations &  0.2488\\
     + augmentations + additional head &  0.2611\\
     + augmentations + additional head + biformer &  0.2680\\
     + augmentations + additional head + biformer + simam &  0.2692\\
     + augmentations + additional head + biformer + simam + flip images &  0.2719\\
     \hline
\end{tabularx}
}

\subsection{\protect\msecond{}{} YOLOv7 Depth - Widened}
\label{usv-dist:depth_widened}
\emph{Matej Fabijani\'{c}, Fausto Ferreira}\\
\emph{University of Zagreb Faculty of Electrical Engineering and Computing}\\
\texttt{\small{\{matej.fabijanic, fausto.ferreira\}}@fer.hr}\\
\\
Our method is based on the YOLOv7 architecture provided by the MaCVi 2025 Distance Estimation Challange organizers, which was already customized to include metric distance estimation for detected objects. Our primary contribution was further enhancing the architecture by increasing its width to improve performance.\\
We increased the model's capacity by modifying the \textit{width\_multiple} parameter, which scaled up the number of feature channels in each layer, leading to a total of approximately 49 million parameters. We utilized the default training configuration provided by the organizers with a few key adjustments. Specifically:
\begin{itemize}
    \item \textbf{Optimization: } The distance loss function was weighted, doubling the loss when the ground-truth object distance was below 441 meters. We have also changed the model fitness calculation to consider only the mAP@50:95 metric as that was the metric scored for the challenge.
    \item \textbf{Augmentation: } We employed HSV (hue-saturation-value) augmentation, left-to-right flipping with a 50\% probability, and increased the scale parameter to 2.0, which significantly improved model performance.
\end{itemize}
Our model was trained solely on the dataset provided by the challenge organizers. The
training set comprised 95\% of the images, with the remaining 5\% reserved for validation. No external datasets or pre-trained weights were used. This resulted in 2793 images used for training and 147 images used for validation. However, we first trained the custom YOLOv7 model until weight convergence without considering the relative distance error, only training for object detection like a standard YOLO. We then used this converged model as the starting point when training for both object detection and metric distance estimation.\\
Training was performed on 2x Nvidia GTX 1080 GPUs. The inference speed of the model
was approximately 0.15 seconds per image.\\
During analysis, we observed challenges such as the difficulty of detecting distant buoys
and markers due to resolution limitations, vessel motion, and adverse visibility conditions. These observations inspired adaptations like the increased scale parameter to focus on the central, more informative parts of the images.\\
Empirical testing revealed that increasing the scale factor provided the largest model
performance boost regarding mAP50-95 metric and relative distance loss metric. This
adaptation aligns with the dataset characteristics, where objects of interest are typically near the image's center.\\
One of the first ideas was to go through the provided dataset and check if the annotations needed corrections to make the bounding boxes fit more "tightly" on the objects, especially the farther ones. However, modifying the annotations and training on "tighter" bounding boxes severely decreased performance on the hidden test dataset, even though the model achieved higher mAP50-95 and lower relative distance error numbers on the modified data. This suggests that adding another expected output to the YOLOv7
architecture might lead to easier overfitting during training.

\subsection{Baseline methods}
\label{usv-eseg:baselines}
\noindent
\emph{MaCVi Organizers}\\

\noindent
We include a slightly adapted method from \cite{kiefer2025approximatesupervisedobjectdistance} as baseline method.

\section{USV-based Obstacle Segmentation}
\label{sec:reports/usvseg}

\subsection{\protect\mfirst{} WaterFormer}

\label{usv-seg:WaterFormer}
\noindent
\emph{Seongju Lee, Junseok Lee, Kyoobin Lee}\\
\texttt{\small{\{lsj2121, junseoklee, kyoobinlee\}@gm.gist.ac.kr,
}}\\
\emph{Artificial Intelligence Laboratory at Gwangju Institute of Science and Technology (GIST)}\\

\begin{figure}[h]
  \includegraphics[width=\columnwidth]{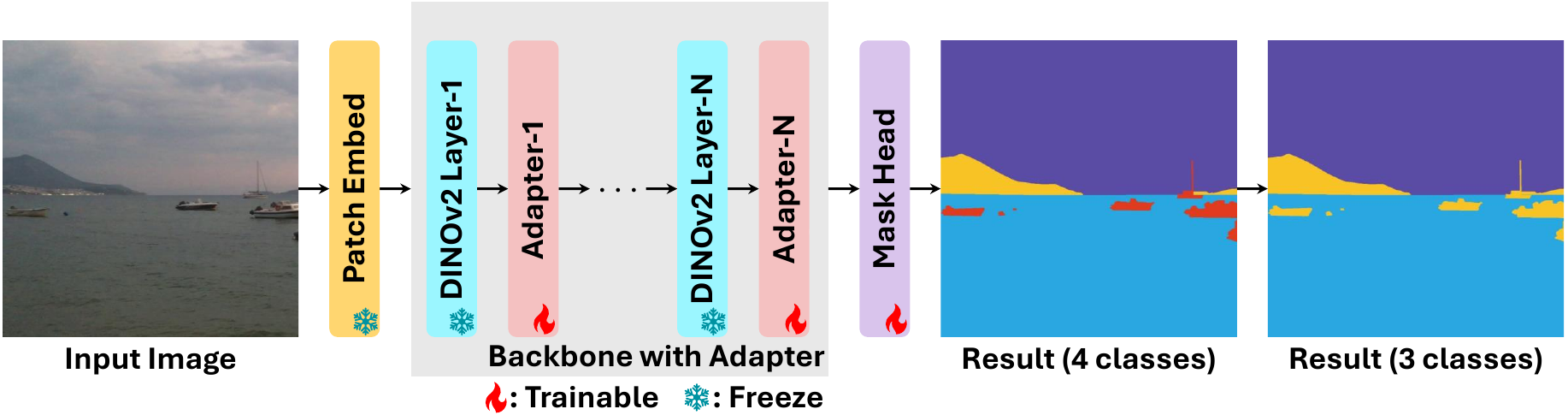}
  \caption{Brief model architecture of WaterFormer}
  \label{fig:model}
\end{figure}

\noindent
\textbf{Method.} We introduce \textbf{\textit{WaterFormer}}, a model for USV-based obstacle segmentation illustrated in Figure \ref{fig:model}, which leverages parameter-efficient fine-tuning of vision foundation models using adapters. Specifically, we used Mask2Former \cite{cheng2021mask2former} with the DINOv2 (Large) \cite{oquabdinov2} backbone. We inserted an adapter with an FC(1024, 128)-GELU-DC-GELU-FC(128, 1024) architecture between the DINOv2 layers, where FC($a$, $b$) is a fully connected layer that takes an $a$-D feature as input and outputs a $b$-D feature, GELU is the Gaussian Error Linear Unit \cite{hendrycks2016gaussian}, and DC is deformable convolution \cite{xiong2024efficient}. During training, the weights of DINOv2 layers are frozen. Our implementation is based on the MMSegmentation framework \cite{segmodelsrepo}.\\

\noindent
\textbf{Training Details.} For training, AdamW \cite{loshchilov2017fixing} optimizer was employed with a learning rate of 0.0001, a weight decay of 0.05, epsilon set to $10^{-8}$, and betas set to (0.9, 0.999). The batch size was set to 8, and the training iterations were set to 40,000. The augmentations included random resizing with scales defined by \verb|[int(512*x*0.1) for x in range(5, 21)]|, using the shortest edge of the image as a standard and a maximum size of 2048. This was followed by random cropping with a defined size of 512 × 512 pixels, random horizontal flipping with a 50\% probability, and photometric distortions to introduce visual variation.\\

\noindent
\textbf{Datasets and Pretrained Weight.} We used only the LaRS \cite{Zust2023LaRS} dataset during training. As described in \cite{oquabdinov2}, the DINOv2 backbone was initialized with a pretrained model, which was trained on the LVD-142M dataset.\\

\noindent
\textbf{Hardware and FPS.} We trained and evaluated \textbf{\textit{WaterFormer}} using an NVIDIA GeForce RTX 4090, achieving an inference speed of 1.7 FPS on this device.\\

\noindent
\textbf{Other Adaptations.} Unlike the original benchmark of LaRS \cite{Zust2023LaRS}, which is a 3-class setup (water, sky and obstacles), we trained \textit{\textbf{WaterFormer}} with a 4-class setup (water, sky, static obstacles, and dynamic obstacles) using panoptic segmentation labels to improve discrimination between dynamic and static obstacles. For the evaluation results, static and dynamic obstacles were merged into a single obstacle class. This decision was inspired by the observation that dynamic obstacles are usually located on the water surface.\\


\subsection{\protect\msecond{} WSFormer}
\label{usv-seg:WSFormer}
\noindent
\emph{Shanliang Yao, Runwei Guan, Xiaoyu Huang, Yi Ni}\\
\texttt{\small{Shanliang.Yao@liverpool.ac.uk, Runwei.Guan@liverpool.ac.uk, x.huang42@liverpool.ac.uk, Yi.Ni21@student.xjtlu.edu.cn,
}}\\
\emph{Yancheng Institute of Technology, Hong Kong University of Science and Technology (Guangzhou), University of Liverpool, Xi’an Jiaotong-Liverpool University}\\

\noindent
\textbf{Method:} We designed a novel marine image semantic segmentation method, named WSFormer, inspired by Mask2Former \cite{cheng2021mask2former}. Mask2Former utilizes masked-attention to focus on local features centered around predicted segments, thereby facilitating faster convergence and enhanced performance. Moreover, it employs multi-scale high resolution to accurately segment small regions. In developing WSFormer, we adopted the Swin-L architecture as the backbone, offering a robust and scalable framework for handling complex water scene data. Additionally, we integrated MultiHeadAttention from the Mask2Former framework, which significantly enriches feature representation and spatial awareness in our segmentation task.\\

\noindent
\textbf{Datasets:} For the pretraining phase, we employed the WaterScenes dataset \cite{yao2024waterscenes}, a comprehensive collection of water-related scenes that spans a diverse array of environments and conditions. This dataset is particularly well-suited for our purposes, owing to its extensive coverage of various water scenarios accompanied by semantic segmentation annotations. Following the pretraining stage, we fine-tuned our model using the LaRS dataset \cite{Zust2023LaRS}, as specified in this challenge.\\

\noindent
\textbf{Training:} We conducted our training on a NVIDIA RTX 4090 GPU, equipped with 24GB of memory. The training process was configured with a batch size of 2 to ensure efficient computation and memory utilization. For optimization, we employed the AdamW optimizer, with a learning rate of 0.0001, an epsilon value of 1e-8 to prevent division by zero, a weight decay of 0.05 for model regularization, and betas set to (0.9, 0.999) to control the exponential moving averages of the gradient and its square. To facilitate a stable and effective learning process, we implemented a warm-up strategy followed by a polynomial learning rate scheduler, spanning a total of 160,000 iterations. This scheduling approach allowed for a gradual increase in the learning rate at the outset, followed by a smooth decay, thereby enhancing the model's convergence and performance.
To enhance the robustness and generalization of our model, we employed a suite of data augmentation techniques, including RandomResize, RandomCrop, RandomFlip, and PhotoMetricDistortion. Additionally, we integrated three loss functions—classification loss, mask loss, and dice loss—to comprehensively measure the discrepancy between predicted and true class labels, ensure accurate segmentation mask prediction, and optimize the model's performance for both classification and segmentation tasks.\\

\noindent
\textbf{Inference:} We utilized the same device for inference as for training. Additionally, we implemented Test Time Augmentation (TTA), a sophisticated data augmentation strategy tailored specifically for the testing phase. By aggregating predictions from each augmented version of the image, we generated a more precise and reliable final prediction, thereby enhancing the overall accuracy and robustness of our model's performance.\\

\noindent
\textbf{Results:} We achieved the 2nd place in the MaCVi 2025 challenge. Our WSFormer model receives an evaluated Quality (Q) value of 79.8, which is a composite metric that integrates general segmentation quality, as measured by the mIoU, and detection quality, as assessed by the F1 score.

\subsection{\protect\mthird{} Advanced K-Net}
\label{usv-seg:Advanced_KNet}
\noindent
\emph{Himanshu Kumar}\\
\texttt{\small{himanshu9082@gmail.com}}\\
\emph{Independent Researcher}\\


\noindent
All of my submissions are marked with Individual as an author in the leaderboard. I have used the K-Net: Towards Unified Image Segmentation method provided in the mmsegmentation-macvi repo.\newline
I have experimented with different Swin transformer backbones, particularly Swin tiny and large. Swin Tiny is trained on the ImageNet-1K dataset. Swin large is pre-trained on a large dataset, ImageNet-22K, which can significantly improve performance on smaller datasets. No extra dataset was used to train the model. Swin large has two variants, with a patch size of 224 and 384. The larger patch variant performed better in terms of mIoU. To capture finer details of obstacles, I varied the image scale up to the size of 2560 x 1280. For optimization, I replaced Adam with Lion, which improves convergence and resulted in a marginal increase in mIoU. I tried introducing stochasticity to the model with a stochastic head, which did not yield significant results. 
\newline For the training pipeline, resizing, random crop and filling, normalization, photometric distribution, and padding were applied. For validation and inference, only Resize and normalization were performed. 
\newline 
The best-performing model in terms of Quality has $swin\_large\_patch4\_window7\_224\_22k\_20220308-d5bdebaf.pth$ as the backbone, and images resized to 2048x1024. I have trained the model on a single Nvidia A100 GPU with a batch size of 4 for 80,000 iterations. AdamW, as an optimizer with a weight decay of 0.0005 and an initial learning rate of 6e-05, was used, and it decayed by 0.1 after 60000 and 72000 iterations. Gradient clipping and StepLR scheduler were incorporated into training. The best validation mIoU model was saved every 8000 iterations. The model took 26 hrs for training, and 2.5 images per second was the inference speed.



\section{USV-based Embedded Obstacle Segmentation}
\label{usv-eseg:submissions}

\subsection{\protect\mfirst{} RSOS-Net}
\label{usv-eseg:rsos-net}
\noindent
\emph{Yuan Feng}\\
\texttt{\small{fengyuan@dlum.edu.cn,  931772830@qq.com}}\\
\emph{Dalian Maritime University, School of Marine Engineering}\\

\noindent\textbf{Method:}\\ Inspired by the encoder-decoder architecture, a realtime surface obstacle segmentation network (RSOS-Net) is created to enable online surface-obstacle detection for an unmanned waterborne vehicle. The network architecture primarily consists of an encoder section and a decoder section. In the encoder, we utilize ResNet-101~\cite{he2016deep} as the backbone network to extract multi-scale features. Additionally, at the final stage of the encoder, we have designed the fast pyramid pooling module (FPPM), which effectively expands the receptive field by combining global average pooling and cascaded average pooling operations, enabling it to capture global and local feature information. It is noteworthy that the integration of local and global contextual information is crucial for distinguishing between water surface disturbances and real obstacles. The decoder comprises a lightweight FPN~\cite{lin2017feature}  and an attentionbased feature fusion module (AFFM). Specifically, RSOSNet utilizes a standard encoder network to generate multiresolution features, which are then seamlessly integrated into the decoder through the FPN structure. Notably, unlike the original FPN, RSOS-Net employs separable convolutions to replace the 3 × 3 convolutions in the original FPN and reduces the number of channels to 256, thereby achieving a lightweight FPN structure. Moreover, the AFFM module, employing a channel-spatial attention mechanism, causes RSOS-Net to focus more on real obstacles, reducing false positives and missed detections. Finally, the outputs of the lightweight FPN are fused through the AFFM and passed through an FCN segmentation head to obtain the final results.

\noindent\textbf{Training:}\\ 
We implemented RSOS-Net using the MMSegmentation framework with an image input size of 768 × 384 and a batch size of 16. The training was conducted on an NVIDIA RTX 3090 GPU equipped with 24G of memory. During the training process, SGD was employed, with a total of 160, 000 iterations conducted. In terms of data augmentation, techniques such as random flipping, cropping, brightness adjustment, and saturation modification were incorporated to enhance the diversity of the training data. As for the dataset, LaRS~\cite{Zust2023LaRS} was exclusively utilized for both the training and validation phases. The pretrained weights of the backbone network were sourced from ImageNet-1K. Regarding the loss function, a combination of cross-entropy loss, dice loss, and focal loss was adopted, with a specific ratio of 1 : 2 : 1.

\noindent\textbf{Observations:}
\begin{itemize}
    \setlength{\itemsep}{1pt}
    \setlength{\parskip}{0pt}
    \setlength{\parsep}{0pt}
    \item In the versions of RSOS-Net we submitted, the one with the ID 14161, which has a Q-score of 64.2 and an FPS of 85.1, uses ResNet-101 as its backbone network. The other version with the ID 14100, featuring a Q-score of 63.6 and an FPS of 102.8, employs ResNet-50 as its backbone. Both versions share the same training methodologies apart from the backbone networks.
    \item The work on RSOS-Net has been summarized into a paper and submitted for publication. If all goes well, it will be publicly available soon. The code for RSOS-Net is publicly accessible at \url{https://github.com/Yuan-Feng1998}.
\end{itemize}

\subsection{\protect\msecond{} Seg-aware Ensemble (EFFNet)}
\label{usv-eseg:effnet}
\noindent
\emph{Yi-Ching Cheng, Tzu-Yu Lin, Chia-Ming Lee, Chih-Chung Hsu}\\
\texttt{\small{\{annie30133, lspss97127, zuw408421476\}@gmail.com, cchsu@gs.ncku.edu.tw
}}\\
\emph{ACVLab, National Cheng Kung University
}\\

\noindent\textbf{Method:}\\
We trained two models and combined them to get the final ensemble model. Inspired by the approach of last year’s second-place team, we selected the lightweight network architectures and further improved prediction accuracy through the ensemble technique. The first model employs ResNet-50 as the encoder paired with DeepLabv3+ as the decoder, while the second uses ResNet-101 with DeepLabv3+. Both backbones were pretrained on ImageNet, and only the LaRS dataset was used for training.

\noindent\textbf{Training:}\\
The training configurations for the two models are outlined below:

\begin{itemize}
    \setlength{\itemsep}{1pt}
    \setlength{\parskip}{0pt}
    \setlength{\parsep}{0pt}
    \item Device: 2 × RTX 2080 Ti / 2 × RTX 3090
    \item Epochs: 150 / 200
    \item Batch size: 16 / 32
    \item Learning rate: 0.0001 for both
    \item Loss function: Dice loss + weighted cross-entropy loss for both
    \item Optimizer: AdamW for both
    \item Scheduler: Cosine annealing for both
    \item Augmentations: Horizontal flip, resize / Horizontal flip, resize, color jitter, unsharp masking, affine transform, and rotation
\end{itemize}

\noindent\textbf{Observations:}
\begin{itemize}
    \item We observed that extending the training duration consistently enhanced model performance, whereas additional augmentations and larger backbones did not yield significant improvements. Using ResNet-50 as the encoder with random horizontal flip augmentation resulted in strong performance, achieving a Q score of 60.8, F1-score of 63.6, and mIoU of 95.6 on the leaderboard.
    \item To capture more complex semantic features, we configured the encoder to output multi-scale feature maps.
    \item In experiments involving different encoders, we noticed substantial performance drops in some models after quantization. Selecting an encoder that aligns with the characteristics of maritime data and maintains stability post-quantization could be a promising direction for further improvement.
\end{itemize}

\subsection{Baseline methods}
\label{usv-eseg:baselines}
\noindent
\emph{MaCVi Organizers}\\

\noindent
We include winning methods from last year's competition eWaSR-RN50 and Mari-MobileSegNet as baseline methods.

\section{USV-based Panoptic Segmentation}
\label{usv-pan:submissions}

\subsection{\protect\mfirst{} MaskDINO}
\label{usv-pan:first}
\noindent
\emph{Jannik Sheikh\textsuperscript{1}, Andreas Michel\textsuperscript{1}, Wolfgang Gross\textsuperscript{1}, \textsuperscript{2}Martin Weinmann}\\
\texttt{\textsuperscript{1}\{firstname.lastname\}@iosb.fraunhofer.de, \textsuperscript{2}martin.weinmann@kit.edu}\\
\emph{Fraunhofer IOSB}\\

\noindent This technical report describes our solutions to the $3^{rd}$ Workshop on Maritime Computer Vision (MaCVi) USV-based Panoptic Segmentation challenge. We employed Mask DINO~\cite{Li2022Mask}, an extension of the detection transformer model DINO \cite{zhang2022dino}. By incorporating a mask prediction branch and creating a pixel embedding map derived from the backbone and transformer encoder features, the model is capable of performing various image segmentation tasks, including panoptic segmentation. We made use of the official implementation of the COCO pre-trained Mask DINO that utilizes an ImageNet-22K \cite{deng2009imagenet} pre-trained Swin-L \cite{liu2021swin} Vision Transformer. Our training was performed on four NVIDIA V100 32GB graphics cards. We trained Mask DINO on the LaRS~\cite{Zust2023LaRS} dataset for 45000 iterations. AdamW \cite{loshchilov2018decoupled} was used as an optimizer with a learning rate of \texttt{1e-4}.

During training, the images were randomly flipped horizontally, resized at multiple scales within a defined range and cropped to 1024×1024 pixels. In addition, gradient clipping and a learning rate multiplier of 0.1 were applied to the backbone. Testing was performed on an NVIDIA V100 32GB graphics card. The inference speed was approximately one FPS. The test pipeline consists of resizing the input image to $1242\times 2208$ ($H\times W$). The proposed model achieves a PQ of 53.9 with an F1-Score of 76.5 on the LaRS test dataset. Further analysis of the results showed a very high $PQ_{st}$ of 92.3 and a significant performance decline of $PQ_{th}$ with 39.4. To address this performance gap, we investigated the potential integration of a Vision Language Model, specifically Llama 3.2 Vision \cite{grattafiori2024llama3herdmodels}. Concretely, for any instance detected by Mask DINO belonging to the "things" category, we assessed whether Llama would provide the same category name as Mask DINO given the ground truth categories. By aligning the results of these two models, we aimed to improve the recognition quality (RQ) score and thus the overall quality. Although the performance (PQ of 52.1) of that combined approach did not improve the individual performance, the results showed promising trends that warrant further research.

\subsection{\protect\msecond{} PanopticWaterFormer}
\label{usv-pan:second}
\noindent
\emph{Seongju Lee, Junseok Lee, Kyoobin Lee}\\
\texttt{\{lsj2121,junseoklee\}@gm.gist.ac.kr, kyoobinlee@gist.ac.kr}\\
\emph{Gwangju Institute of Science and Technology (GIST)}\\

\paragraph{Method.}
\textit{PanopticWaterFormer} combines the inference results of WaterFormer for semantic segmentation and Cascade Mask R-CNN~\cite{cai2019cascade} with the InternImage (Large)~\cite{wang2023internimage} backbone for instance segmentation. This approach harnesses the strong segmentation capability of WaterFormer for stuff classes. The detailed model architecture of WaterFormer is described in the USV-based Semantic Segmentation section. Our implementation is based on MMDetection \cite{chen2019mmdetection} framework.

\paragraph{Training Details.}
For training Cascade Mask R-CNN, the AdamW~\cite{loshchilov2017fixing} optimizer was employed with a learning rate of 0.0001, a weight decay of 0.05, epsilon set to $10^{-8}$, and betas set to (0.9, 0.999). The batch size was set to 2, and the number of training epochs was set to 24. The learning rate was adjusted at epochs 8 and 11 by multiplying it by a factor of 0.1, with updates applied at the epoch level. Additionally, the layer-wise learning rate decay was set to 0.9 to preserve the pretrained knowledge in the earlier layers. The augmentation pipeline for training includes random horizontal flipping with a probability of 50\%, large-scale jittering \cite{ghiasi2021simple} through random resizing with a scale range of (0.1, 2.0) while maintaining the aspect ratio, and random cropping to a fixed size of 1024$\times$1024. During inference, the images were resized to a fixed size of 2048$\times$1024.

\paragraph{Datasets and Pretrained Weights.}
We used only the LaRS~\cite{Zust2023LaRS} dataset during training. To initialize Cascade Mask R-CNN with the InternImage backbone, we used COCO-pretrained weights~\cite{coco} at the start of training.

\paragraph{Hardware and FPS.}
We evaluated \textbf{\textit{PanopticWaterFormer}} using an NVIDIA H100, achieving an inference speed of 1.1 FPS on this device.

\paragraph{Panoptic Mask Generation.}
First, we obtained a segmentation map from WaterFormer, which divides the scene into stuff (water, sky, and obstacle) and things (dynamic obstacles). Second, we extracted instance masks, their corresponding bounding boxes, and predicted labels from Cascade Mask R-CNN. Instances with a confidence score below 0.5 were discarded. Finally, panoptic segmentation maps were generated by assigning instance masks to the things region of the segmentation map in descending order of size, from largest to smallest, if their bounding boxes overlapped with the things region.

\subsection{\protect\mthird{} FC-CLIP-LaRS}
\label{usv-pan:third}
\noindent
\emph{Josip \v{S}ari\'c}\\
\texttt{josip.saric@fri.uni-lj.si}\\
\emph{University of Ljubljana}\\

\noindent Our submission builds upon the 
panoptic mask transformer FC-CLIP~\cite{yu23neurips}.
FC-CLIP incorporates a
language-based classifier, 
which enables open-vocabulary recognition~\cite{radford21icml}. 
Specifically, the classification scores
are derived from similarities
between the predicted mask embeddings
and the embeddings generated by
a frozen CLIP~\cite{radford21icml} 
text encoder
applied to class descriptions.
To enhance the recognition
of unseen classes,
the scores are ensembled 
with zero-shot classification
of mask-pooled features~\cite{ghiasi22eccv,yu23neurips}
from frozen convolutional 
CLIP visual encoder~\cite{radford21icml}.
Note that the same module
serves as an image feature extractor
for mask decoder.
Concretely, we rely on
ConvNext-L~\cite{liu22cvpr} backbone
initalized with OpenCLIP weights~\cite{cherti23cvpr}
pretrained on the LAION dataset~\cite{schuhmann22neurips}.



Our preliminary study showed that a pretrained COCO~\cite{coco} FC-CLIP results in a poor performance (18.2 PQ and 50.1 mIoU) on the LaRS validation set~\cite{Zust2023LaRS}.
We hypothesize this is due to a significant domain shift between COCO and LaRS.
We also noticed that the default LaRS class text descriptions 
are short and vague, which could affect the performance
of the language-based classifier.
To test this hypothesis, we extended the class descriptions
with additional synonyms or hyponyms
(e.g.\ \emph{static-obstacle} with 
\emph{building}, \emph{terrain}, \emph{vegetation}, etc.). This lead to substantial performance boost (27.6 PQ and 91.6 mIoU).


Finally, we fine-tuned
the COCO model on LaRS train
for 90k iterations 
with batch size 8 
and base learning rate $10^{-4}$.
We set the crop size 
to $512 \times 1024$ pixels,
and applied random color
and scale jitter.
During evaluation 
we resize the larger side
to 2048 pixels,
while maintaining
the original aspect ratio.
The fine-tuned model achieves 
53.6 PQ and 94.9 mIoU 
on the LaRS validation set. 
On the test set, 
the model achieved 49.7 PQ, 
securing the third place in the 
official MaCVi Panoptic Segmentation Challenge.

It is important to note that 
the results of the fine-tuned model 
do not reflect its open-vocabulary capabilities, 
as the training and testing vocabularies are identical. 
In future work, we aim to explore 
open-vocabulary panoptic segmentation 
specifically for the maritime domain.

\subsection{Baseline and PanSR}
\label{usv-pan:baselines}
\noindent
\emph{MaCVi Organizers}\\

\noindent
As the baseline, we train the Mask2Former~\cite{cheng2021mask2former} with the Swin-B backbone using MMSegmentation~\cite{mmseg2020} framework with a training configuration adapted for LaRS~\cite{Zust2023LaRS}. Furthermore, we also submitted our novel PanSR method~\cite{Zust2024PanSR} to the leaderboard. PanSR introduces a new approach to query proposal generation and rethinks the training methodology for better synergy with the proposal-based queries. This leads to more robust detection of small objects and state-of-the-art performance on LaRS. More details are available in \cite{Zust2024PanSR}.

\section{Underwater Image Restoration}
\label{uir:submissions}
\subsection{\protect\mfirst{} Team NJUST-KMG}
\label{uir:first}
\noindent
\emph{Yipeng LIN$^1$, Xiang yang$^1$, Nan Jiang$^2$, Yang Yang$^1$}\\
\texttt{2396599652@qq.com, maybe\_1221@163.com, 1306550598@qq.com, yyang@njust.edu.cn}\\
\emph{$^1$Nanjing University of Science and Technology}\\
\emph{$^2$Nanjing University}\\
\begin{figure}[!hbtp]
    \centering
    \includegraphics[width=1\linewidth]{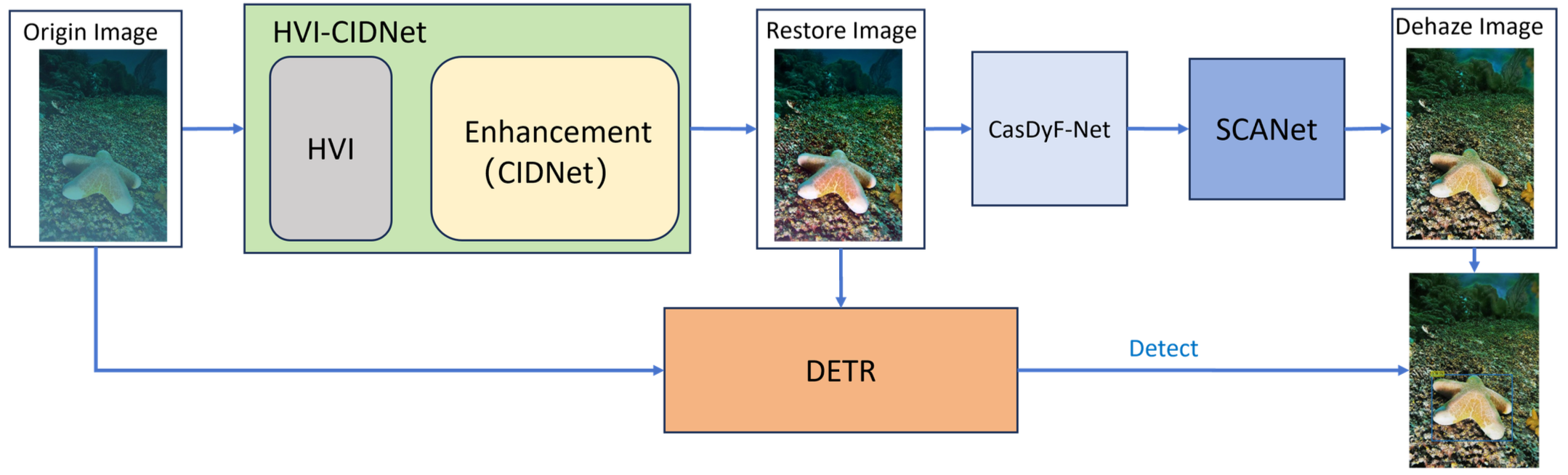}
    \caption{Overview of the architecture proposed by Team NJUST-KMG for underwater image restoration and marine species detection.}
    \label{fig:Team1_NJUST}
\end{figure}
The overall framework of our method is illustrated in Figure \ref{fig:Team1_NJUST}. The challenge is divided into two parts: image restoration and object detection. Due to the lack of an end-to-end model framework, we trained separate models for these two parts. For the image restoration task, we adopted a two-stage approach. Since the dataset primarily consists of large images, we used the efficient image restoration method HVI-CIDNet \cite{yan2024you} in the first stage. In the second stage, we employed the image dehazing methods CasDyF-Net \cite{yinglong2024casdyf} and SCANet \cite{guo2023scanet}. For object detection, we used the DETR \cite{carion2020end} model. With the rapid development of deep learning, many powerful deep learning models \cite{wan2024covlr, xu2024itp, yang2024facilitating}, have emerged. We experimented with several image restoration models, including U-shape \cite{peng2023u}, UDAformer \cite{shen2023udaformer}, and the solution from team SYSU-FVL-T2 for the NTIRE 2024 Low Light Enhancement Challenge \cite{liu2024ntire}. However, we found that due to the large images in the RSUIGM \cite{rsuigm} dataset, these models performed poorly during training. Therefore, we used HVICIDNet, an efficient and outstanding image restoration network. However, upon observing the restored images, we noticed large areas with shallow color overlays that resembled artificially generated fog. Therefore, after the first stage, we restored the training dataset and used them to train dehaze models CasDyF-Net and SCANet. The restored images were first processed by the fine-tuned CasDyF-Net and then by the pre-trained SCANet.For the object detection task, we used DETR as our main model and adopted a two-stage fine-tuning strategy. The model was trained on the original dataset and then fine-tuned on the dataset with initially restored images. In order to mitigate the impact of the long-tail distribution, we applied weight allocation to the classification head. Finally, we achieved the following results: for image restoration, CCF: 0.077, UCIQE: 0.571, and UIQM: 0.7; and for object detection, mAP: 0.125, F1: 0.125, and mIoU: 0.693. And our model can process two large images similar size to test dataset per second on an RTX 3090.
\subsubsection{Training Details}
Dataset. We only used the official RSUIGM dataset for training. Additionally, the pre-trained models HVI-CIDNet, CasDyF-Net, SCANet and DETR are pretrained on the LOLV1 \cite{wei2018deep}, Haze4K \cite{liu2021synthetic}, NTIRE and ImageNet \cite{russakovsky2015imagenet}, respectively.
\begin{table}[h!]
    \centering
    \resizebox{\linewidth}{!}{
    \begin{tabular}{l c c c c c c}
        \hline
        \textbf{Method} & \textbf{mAP} & \textbf{F1} & \textbf{mIoU} & \textbf{CCF} & \textbf{UCIQE} & \textbf{UIQM} \\ \hline
        baseline & 0 & 0 & 0.00 & 0.04 & 0.498 & 0.193 \\
        HVI-CIDNet+DETR & 0.093 & 0.093 & 0.691 & 0.038 & 0.547 & 0.493 \\
        HVI-CIDNet+DETR(LT) & \textbf{0.143} & \textbf{0.143} & \textbf{0.636} & 0.038 & 0.547 & 0.493 \\
        HVI-CIDNet+CasDyF-Net+DETR(LT) & 0.090 & 0.090 & 0.561 & 0.032 & 0.530 & 0.696 \\
        HVI-CIDNet+CasDyF-Net+SCA-Net+DETR(LT) & 0.125 & 0.125 & 0.693 & 0.077 & \textbf{0.571} & \textbf{0.700} \\ \hline
    \end{tabular}
    }
    \caption{Performance comparison of different methods}
    \label{tab:team1_mvrc}
\end{table}

Implementation Details.We are training models on a single RTX 3090. During the training process, the dataset is split into a training set and validation set by 9:1. For HVI-CIDNet, the training strategy from MIRNet-v2 \cite{waqas2022learning} is adopted, where images are progressively trained at sizes of 256/512/1024 for 30/30/10 epochs, respectively, and save the highest PSNR model. For CasDyF-Net, the repaired dataset is used for 10 epochs of training. For DETR, we fine-tuned the pre-trained DETR model (with a ResNet50 backbone) for 200 epochs on the initial dataset, and then fine-tuned it for 50 epochs on the restored images. The experimental results are shown in the Table \ref{tab:team1_mvrc}.

\subsection{\protect\msecond{} Team BUPT MCPRL}
\label{uir:second}
\noindent
\emph{Yutang Lu, Fei Feng}\\
\texttt{luyutang8610@163.com, wusaqi09@gmail.com}\\
\emph{Beijing University of Posts and Telecommunications, Beijing, China}\\
\begin{figure}[!hbpt]
    \centering
    \includegraphics[width=1\linewidth]{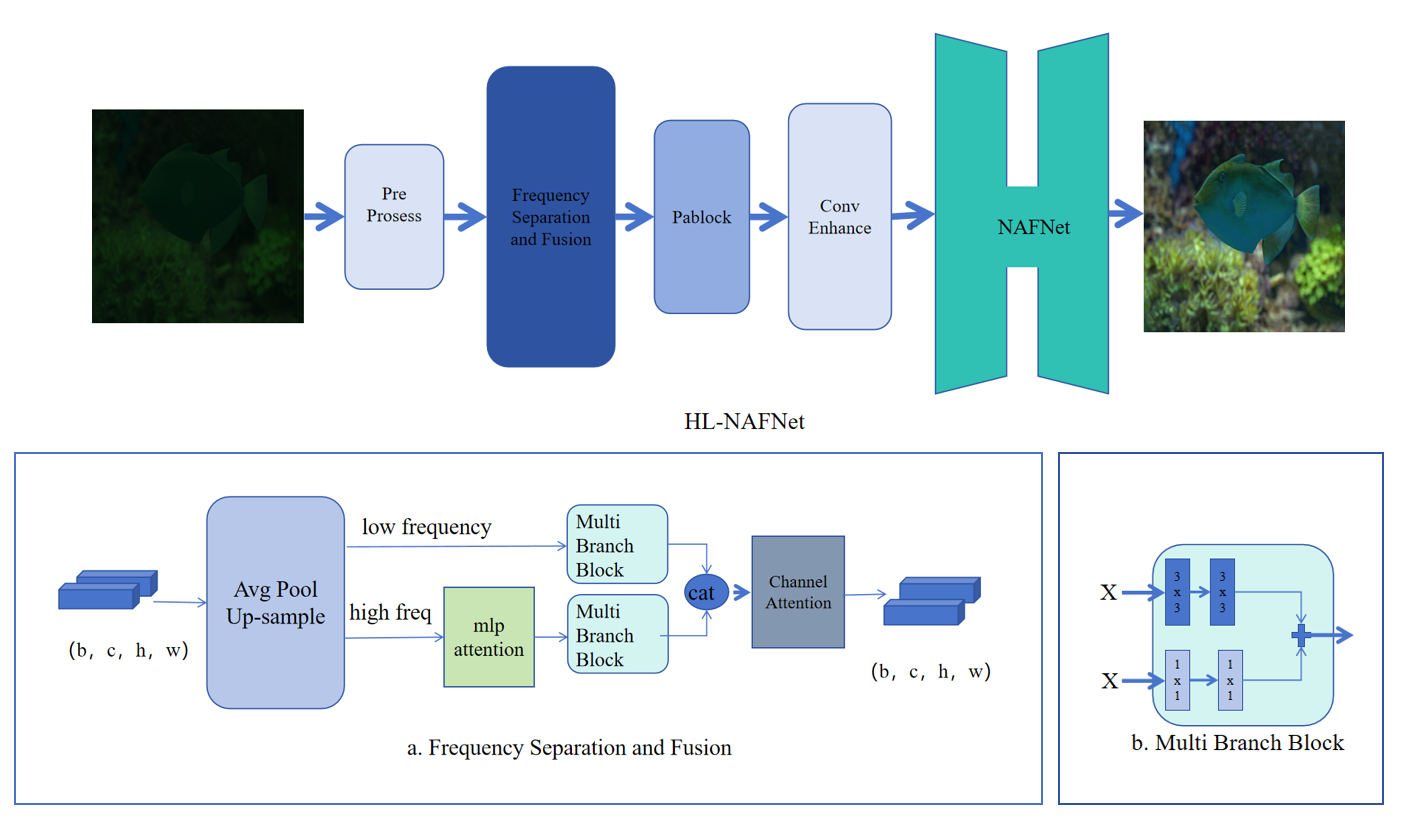}
    \caption{Overview of the architecture proposed by Team BUPT MCPRL for underwater image restoration and marine species detection.}
    \label{fig:hlnafnet}
\end{figure}
 To enhance the visual quality of underwater images, we have proposed the HL-NAFNet model, which integrates frequency separation and fusion technology with NAFNet~\cite{chen2022simple} post-processing. This allows the image to undergo an initial restoration phase, followed by a second phase of NAFNet restoration, aiming to enhance the clarity and color fidelity of the image. The model architecture, as shown in Figure~\ref{fig:hlnafnet}, consists of the following key modules:

\textbf{Frequency Separation and Fusion:} After convolutional pre-processing, we reduce the spatial dimensions of the image through average pooling to extract low-frequency information. High-frequency information is obtained by subtracting the upsampled low-frequency information from the original image. For the high-frequency information, we use Multi-Layer Perceptron Attention to highlight image details. The low and high-frequency feature images are then processed in parallel through a multi-branch block to enhance the expression of features. Finally, channel attention is used to fuse low and high-frequency information, enhancing the final output.

\textbf{PABlock (Pixel Attention Block): }Utilizing Global Average Pooling, this block captures a comprehensive statistical profile of the feature map. It then utilizes 1x1 convolutions with a ReLU activation to craft an attention map.

\textbf{Training Details.}Adopting AdamW optimizer with $\beta_1 = 0.9$ and $\beta_2 = 0.999$ for training 100epochs. The initial learning rate was set to $10^{-4}$. The cosine annealing strategy is employed to decrease the learning rates to $10^{-6}$  . All experiments are conducted on two NVIDIA GeForce RTX 4090 24GB GPUs.in the NAFnet the  enc\_blks is $[2, 2, 4, 8]$, the middle\_blk\_num is $12$ and the dec\_blks is $[2, 2, 2, 2]$.
\subsubsection{detect and classify}
Detecting the restored underwater images based on YOLOv5s~\cite{Jocher_Ultralytics_YOLO_2023}.The data set uses the original underwater images in RSUIGM to filter the original images according to the requirements of target detection and input training. The experimental platform environment is Linux system with Tesla T4 graphics card. The input size is $(1080,1080)$, the batchsize is $32$, the initial learning rate is $0.01$, the SGD optimizer is used, and the epoch is $100$.

\subsection{\protect\mthird{} Team MTU}
\label{uir:third}
\noindent
\emph{Ali Awad$^1$, Evan Lucas$^2$, Ashraf Saleem$^1$}\\
\texttt{aawad@mtu.edu, eglucas@mtu.edu, ashraf@mtu.edu}\\
\emph{$^1$Department of Applied Computing, College of Computing, Michigan Technological University, Houghton, MI 49931, USA}\\
\emph{$^2$Institute of Computing and Cybersystems, Michigan Technological University, Houghton, MI 49931, USA}\\
Our team used the Texture Enhancement Model Based on Blurriness and Color Fusion (TEBCF) method \cite{yuan2021tebcf} based on extensive benchmarking studies that consider the impact of image enhancement on detection in underwater conditions \cite{saleem2025understanding, awad2024evaluatingimpactunderwaterimage} using publicly available object detection datasets, including one made by our group \cite{saleem2023multi,fu2023rethinking}. TEBCF uses multi-scale fusion to combine the outputs of two input pipelines, one that uses the RGB color space to improve sharpness through a dehazing based on a dark channel prior, and another, and another that uses CIELAB space for color correction and brightness adjustment. Since TEBCF is a non-data-driven method, it requires no training and can be used directly for inference on a CPU. The default parameters of TEBCF provided by the authors are used in this work without any modification. We ran TEBCF on a Xeon W-2295 CPU in parallel mode with 18 workers resulting in approximately 4 images per minute. On the detection side, we used the SuperGradients \cite{supergradients} library to implement YOLO-NAS detection model on a Linux server with two Nvidia Tesla V100 GPUs. The training was performed with a batch size of 16, AdamW optimizer, decay of 0.00001, large YOLO architecture, and COCO pre-trained weights. Separate detection models of YOLO-NAS are trained on original and enhanced images. Our findings reveal that image enhancement has the potential to improve object detection in some cases despite its negative overall performance on detection. In addition, we noticed that the current Image Quality Metrics (IQM) are very sensitive and do not accurately represent human perception. For example, TEBCF outperforms many state-of-the-art enhancement models by a large margin in terms of IQM, yet it produces very noisy and reddish images compared to other models. Finally, we noticed that enhancement could help in revealing hidden objects that are missed by human annotators resulting in inaccurate dataset annotations and unreliable results.

\subsection{Team ACVLab}
\label{uir:fourth}
\noindent
\emph{Ching-Heng Cheng, Yu-Fan Lin, Tzu-Yu Lin, Chih-Chung Hsu$^*$}\\
\texttt{henry918888@gmail.com, aas12as12as12tw@gmail.com, lspss97127@gmail.com, cchsu@gs.ncku.edu.tw}\\
\emph{ACVLab, National Cheng Kung University}\\
\emph{$^*$corresponding author}
\subsubsection{Description}
We trained two models to perform inference sequentially to obtain the final result. Drawing inspiration from the color space fusion approach, which is commonly used in UIE tasks, as well as from the attention mechanisms prevalent in natural image restoration (IR), we selected TCTL-Net \cite{li2023tctl}, a network structure based on the LAB color space, as our baseline model. We further modified the original architecture based on our expertise to enhance the restoration quality. Additionally, we added an MLP head to predict the downwelling depth (d) and the attenuation coefficient (k) of a degraded image.
However, the large size of the input image presents a significant challenge to general models, particularly those that rely on global feature extraction. These models are not well-suited for slicing the image into patches to reduce memory usage. To address this, we trained a lightweight fusion model that combines the degraded original-size image with the restored small-size image, resulting in a full-size restored output. Both models were trained using only the provided RSUIGM dataset \cite{rsuigm}.
For the detection component, we utilized Grounding DINO \cite{zhang2022dino} with pre-trained weights as the detector and fine-tuned the threshold settings to achieve improved results.
\subsubsection{Training Details}
The training configurations for the two models are outlined below:
\begin{itemize}
    \item Device: 2 × RTX 2080 Ti / 1 × RTX 2080 Ti
    \item Epochs: 100 / 50
    \item Batch size: 24 / 1
    \item Learning rate: 0.0001 for both
    \item Loss function: L1 loss + L2 loss + SSIM loss/ L1 loss + L2 loss
    \item Optimizer: AdamW for both
    \item Scheduler: Cosine annealing for both
    \item Augmentations: Horizontal flip for both
    \item Inference time: 266ms per image on a 2080Ti GPU
\end{itemize}

\subsubsection{Observations}
We observed that synthetic degraded images with high levels of depth or attenuation may introduce artifacts between aquatic animals and the background. These synthetic degraded images appear unnatural, which causes the restoration results to be dependent on the specific degradation conditions. Furthermore, since the evaluation metrics are non-reference based and our training strategy is based on supervision, the results may not fully capture a rich color representation. This is because the ground truth images may not always be as colorful. This insufficiency represents an area that deserves attention and improvement in our future work.

{\small
\bibliographystyle{ieee_fullname}
\bibliography{egbib}
}

\end{document}